\documentclass[journal,twoside,web]{ieeecolor2}
\usepackage{generic}
\usepackage{cite}
\usepackage{amsmath,amssymb,amsfonts}
\usepackage{algorithmic}
\usepackage{graphicx}
\usepackage{textcomp}
\usepackage{subfiles} 
\usepackage{amsmath,amsfonts}
\usepackage{algorithmic}
\usepackage{algorithm}
\usepackage{array}
\usepackage{subfigure}
\usepackage{textcomp}
\usepackage{stfloats}
\usepackage{url}
\usepackage{verbatim}
\usepackage{graphicx}
\usepackage{cite}
\usepackage{tikz}
\usepackage{mathdots}
\usepackage{yhmath}
\usepackage{cancel}
\usepackage{color}
\usepackage{siunitx}
\usepackage{array}
\usepackage{multirow}
\usepackage{amssymb}
\usepackage{gensymb}
\usepackage{tabularx}
\usepackage{extarrows}
\usepackage{booktabs}
\usetikzlibrary{fadings}
\usetikzlibrary{patterns}
\usetikzlibrary{shadows.blur}
\usetikzlibrary{shapes}

\usepackage{booktabs}
\usepackage{newtxtext} % To fix textsc and textit incompatibilities
\def\BibTeX{{\rm B\kern-.05em{\sc i\kern-.025em b}\kern-.08em
    T\kern-.1667em\lower.7ex\hbox{E}\kern-.125emX}}
\begin{document}
%\title{When Exoskeletons Take Over: User Experience of Rigid Guidance Across Personalized Gait Patterns, a Pilot Study}

\title{The Impact of Gait Pattern Personalization on the Perception of Rigid Robotic Guidance: A Pilot User Experience Evaluation}

\author{Beatrice Luciani$^{1}$, Katherine Lin Poggensee$^{1,2}$, Heike Vallery$^{3,4}$, Alex van den Berg$^{1}$,\\ Severin David Woernle$^{1}$, Mostafa Mogharabi$^{1,5}$, Stefano Dalla Gasperina$^{1,6}$, and Laura Marchal-Crespo$^{1,2}$
% <-this % stops a space
\thanks{$^{1}$Cognitive Robotics Department, Delft University of Technology, Delft, The Netherlands}
\thanks{$^{2}$Department of Rehabilitation Medicine, Erasmus Medical Centre, Rotterdam, The Netherlands.
}
\thanks{$^{3}$Department of Biomechanical Engineering, Delft University of Technology, Delft, The Netherlands.}
\thanks{$^{4}$Institute of Automatic Control, RWTH Aachen University, Aachen, Germany.}
\thanks{$^{5}$Department of Mechanical Engineering, FUM Center of Advanced Rehabilitation and Robotics Research (FUM CARE), Ferdowsi University of Mashhad, Mashhad, Iran.}
\thanks{$^{6}$Walker Department of Mechanical Engineering, University of Texas at Austin, Austin, TX, USA.}
\thanks{Data, code, and Supplementary Materials will be available on Zenodo after publication.}
% <-this % stops a space
}

\maketitle

\begin{abstract}
\textit{Objective:} Exoskeletons modulate human movement across diverse applications, from performance augmentation to daily-life assistance. These systems often enforce specific kinematic patterns to mitigate injury risks and motivate users to keep moving despite diminished capacity. However, little is known about users' perception of such robot-imposed guidance, especially when personalized to the uniqueness of individual human walk. Given the usually substantial computational cost for personalization, understanding its subjective impact is essential to justify its implementation over standard patterns.
\textit{Methods:}  %, training regression models on a normative gait dataset. 
Ten unimpaired participants completed a within‑subject experiment in a multi‑planar treadmill‑based exoskeleton that enforced three different gait patterns: personalized, standard, and a randomly selected pattern from a publicly available database. Personalization was achieved using a data‑driven framework that predicts hip, knee, and pelvis trajectories from walking speed, anthropometric, and demographic data. The standard pattern was obtained by averaging gait patterns from the aforementioned database. After each condition, participants rated enjoyment, comfort, and perceived naturalness. Knee joint interaction forces were also recorded.
\textit{Results:} Subjective ratings revealed no significant differences among patterns, despite all trajectories being executed with high accuracy. However, gait patterns experienced last were rated as significantly more comfortable and natural, indicating adaptation to the system. Higher interaction forces were observed only for the random vs. standard pattern.
\textit{Conclusions:} Personalizing gait kinematics had minimal short‑term influence on user experience relative to the dominant effect of adaptation to the exoskeleton.
\textit{Significance:} These findings highlight the importance of integrating subjective feedback and accounting for user adaptation when designing personalized robot controllers.
\end{abstract}


% Adding a line so that the reviewing/editing button isn't covering text
\section{Introduction}
\IEEEPARstart{E}{xoskeletons} capable of physically guiding human movement are increasingly found in different domains, from augmenting physical performance in sports~\cite{Tan_Collins_2025} to assisting~\cite{vassallo2020} and rehabilitating~\cite{Marchal-Crespo2022} individuals with sensorimotor impairments. 
A common strategy across these applications consists of prescribing predefined joint trajectories to guide the user’s limbs in the ``correct" kinematic pattern~\cite{de2023control, luo_trajectory_2022}. While the strength the robot enforces the desired motion can be modulated, e.g., by adjusting stiffness depending on the user's performance~\cite{agree}, a rigid guidance is still needed to prevent the risk of injuries, especially in those with a very low skill level on the task at hand~\cite{Basalp2021} and people with severe sensorimotor impairment~\cite{de2023control}, while motivating them to keep moving, despite their limitations~\cite{Reinkensmeyer2007}. This is particularly important for gait exoskeletons, as safety is paramount to maintain postural stability and prevent falls. 
In the context of gait exoskeletons, trajectory-based controllers commonly enforce predefined gait trajectories based on pre-recorded data (e.g.,~\cite{Yeung2017}), 
which rarely align perfectly with users' natural gait. In fact, human gait is inherently individual-specific \cite{kale2004} and shaped by factors such as age, gender, body measurements~\cite{ren2019}, as well as gait speed~\cite{hanlon2006}. 
This has motivated the recent development of several methods for gait‑pattern personalization~\cite{Carvalho2025,challa2022}. Model-based methods, for example, generate gait patterns using mathematical~\cite{kagawa2009} or biomechanical models~\cite{falisse2019,hu2022}, incorporating parameters like joint angles, muscle forces, and body dynamics. A major drawback of these approaches is their reliance on complex formulations and assumptions about body dynamics and muscle activation that may not fully capture the complexity of human gait. This limitation can be mitigated by learning-based methods, which derive gait patterns directly from data, allowing the representation of nonlinearities and inter-individual differences that are difficult---or even impossible---to encode analytically. This last class of approaches is becoming increasingly popular due to their proven effectiveness in predicting individualized gait patterns~\cite{ren2019,zou2021}, at the cost of heavily depending on large, high-quality training datasets~\cite{geron2019}. These considerations highlight the need for approaches integrating biomechanical interpretability, capturing how factors such as walking speed and morphology shape kinematics, with the ability to represent inter‑individual variability, without requiring the large datasets typically needed by complex learning‑based models.
 
Despite these advances in generating personalized trajectories, little is known about how robot‑imposed patterns---personalized or not---are actually experienced by users. Prior studies emphasize the importance of integrating users' preferences and subjective experience into the design of robotic devices and guiding  strategies~\cite{ingraham_leveraging_2023}. Yet, the few subjective assessments reported in lower-limb exoskeleton research typically rely on broad usability tools such as System Usability Scale (SUS) or Technology Acceptance Model (TAM) scores, or generic comfort ratings~\cite{zhou_user_2024}. As a result, there is a lack of empirical evidence on whether personalized gait trajectories genuinely improve users’ subjective experience, including factors such as comfort, naturalness, and enjoyment, compared to non-personalized trajectories. Moreover, the field lacks systematic investigations into how unimpaired users, who might be more sensitive to rigid guidance because it conflicts with their own highly optimized natural gait, respond to walking along robot‑imposed personalized versus non‑personalized paths.

% Adding a line break for readability
Driven by these gaps, we conducted a pilot study with ten unimpaired participants to compare their subjective perception in response to walking in exoskeleton-enforced personalized and non-personalized gait patterns. We assessed participants’ perception in terms of \textit{enjoyment}, as it reflects users' engagement and willingness to interact with the system~\cite{papp-schmitt_embodied_2024}; \textit{comfort}, since physical strain during exoskeleton use can impact usability and acceptance~\cite{Wang2018}; and \textit{naturalness}, as prior work has shown that exoskeleton-guidance is often evaluated on its resemblance to human physiological movements, commonly assessed through indicators of movement quality such as smoothness~\cite{Luciani2023}, human-robot coordination~\cite{shushtari_humanexoskeleton_2024}, and high correlation between assisted and unassisted kinematics~\cite{moscatelli_impact_2025}.
We enforced the different gait patterns using an in-house modified Lokomat\textsuperscript{\textsuperscript{\textregistered}} exoskeleton. This treadmill-based exoskeleton can support the already integrated knee flexion/extension in the commercial version, as well as newly actuated hip flexion/extension and ab-/adduction together with full translational and rotational pelvis movements, with actuated pelvis lateral movements. 
The personalized gait patterns, i.e., hip, knee, and lateral pelvis desired trajectories, were generated for each participant using gait pattern prediction models that incorporate walking speed as well as anthropometric and demographic data. These models were trained on a publicly available walking database from 42 healthy people. The trajectories were then enforced using a stiff position-derivative controller. We expected healthy users to prefer their personalized gait patterns in terms of enjoyment, comfort, and perceived naturalness, as they might be specially sensitive to rigid guidance that conflicts with their own natural gait.

\section{Methods}
\subsection{The treadmill-based lower-limb exoskeleton}
\label{Sec:loko}
\begin{figure}[ht!]
    \centering
    \includegraphics[width=\linewidth]{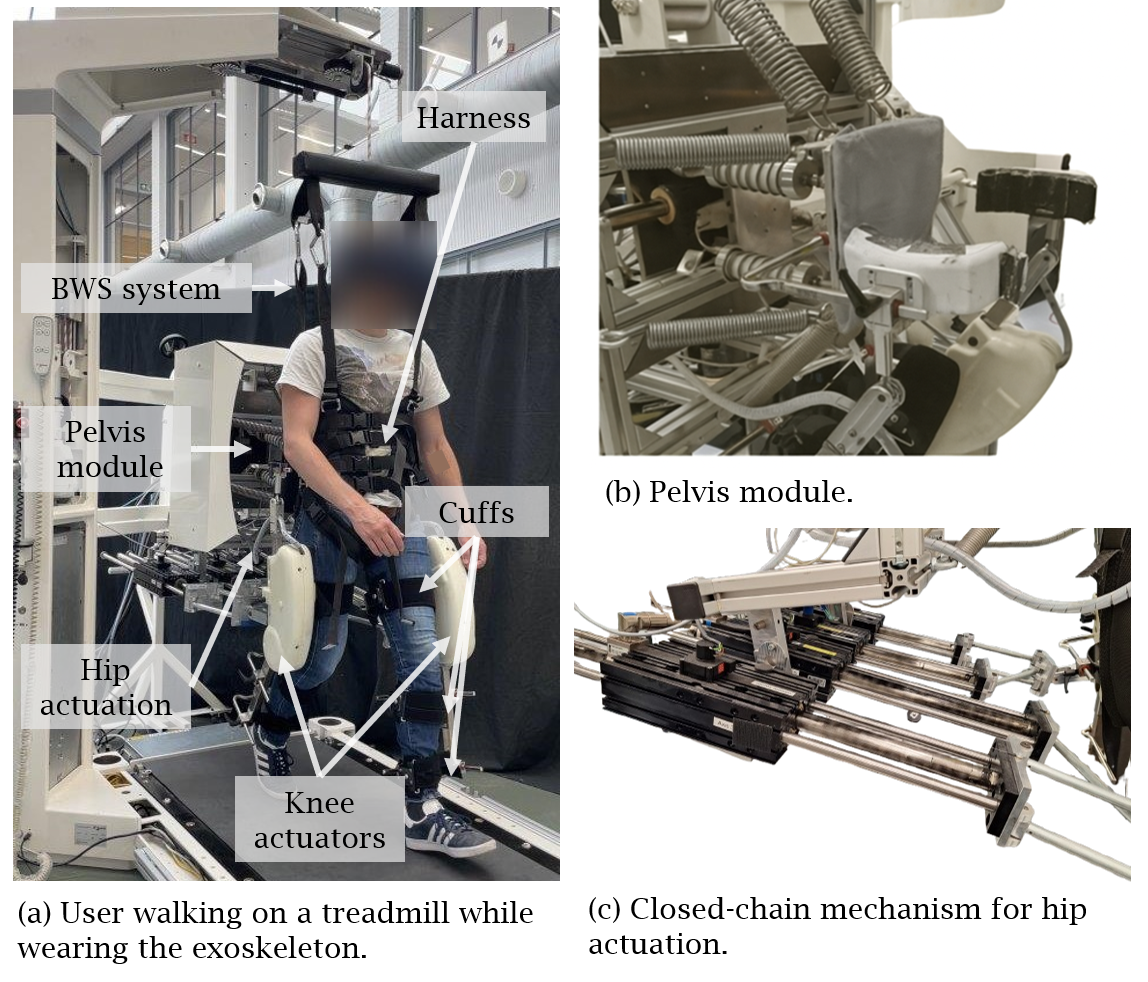}
    \caption{(a) The new exoskeleton with a user secured, featuring an active body weight support (BWS) system with actuated lateral movement capability, a 6-DoF series elastic actuation (SEA) pelvis module with additional lateral translation actuator, two active 2-DoF closed-chain hip actuators, ball-screw-driven 1-DoF knee actuators, ankle/shank/thigh cuffs, and a safety harness for fall prevention. (b) Detail of the 6-DoF pelvis module. (c) Detail of the closed-chain mechanism for hip actuation.}
    \label{fig:experimental_setup}
\end{figure}

For this work, we used a grounded exoskeleton developed at the Sensory Motor Systems (SMS) Lab at ETH Zurich in collaboration with Hocoma AG, Switzerland (Fig.~\ref{fig:experimental_setup}). The new exoskeleton is a modified version of the company's well-known Lokomat\textsuperscript{\textregistered}. In particular,
the system maintains the original Lokomat\textsuperscript{\textregistered}, parts of the body-weight support (BWS) system, the treadmill, and the knee joint flexion/extension ball-screw-driven linear actuation. The system also maintains the in-line force sensor at the knee actuation to measure the human-robot force interaction\cite{lokomatPro2023}. 

The mechanical innovations of the modified exoskeleton include: i) changes in the hip actuation to allow for hip flexion/extension and ab-/adduction~\cite{wyss2019}, ii) a pelvis module that embodies a six degree of freedom (DoF) compliant mechanism that accommodates weight shifting during walking, with one actuated DoF that assists the pelvis lateral movements~\cite{wyss2018multidimensional}, and iii) the original BWS system augmented by one actuated DoF that enables lateral movements to follow the pelvis lateral movements~\cite{wyss2014body}.  
These innovations resulted in a human–exoskeleton interface that incorporates twelve degrees of freedom (DoFs): six at the pelvis, two at each hip, and one at each knee. Among these, seven DoFs are actuated: one at the pelvis (coupled with the new actuated BWS lateral mechanism), two at each hip (flexion/extension and ab-/adduction), and one at each knee. The kinematic model of the exoskeleton is described in detail in the \textit{Supplementary Materials}. Below we summarize the most relevant robot features.

%%%%
The \textbf{pelvis module} includes a pelvis plate, which is attached to the user's pelvis and can undergo both translational and rotational motion thanks to a compliant spring-based connection between the fixed back of the pelvis module and the pelvis plate (Fig.~\ref{fig:experimental_setup}.b; see~\cite{wyss2019} for further details). The pelvis module functions as a series elastic actuator (SEA), with one actuated degree of freedom (P01-48x240 motor, NTI AG LinMot, Switzerland), enabling lateral movement relative to a fixed frame. To prevent hindering the natural lateral movement of the pelvis, the BWS is laterally actuated through a lead-screw mechanism parallel to the pelvis's actuated degree of freedom. Note that the other parts of the BWS remain unchanged with respect to their commercial version, i.e., allow for static and dynamic users' weight unloading. 

\begin{figure}[t!]
    \centering
    \includegraphics[width=\columnwidth]{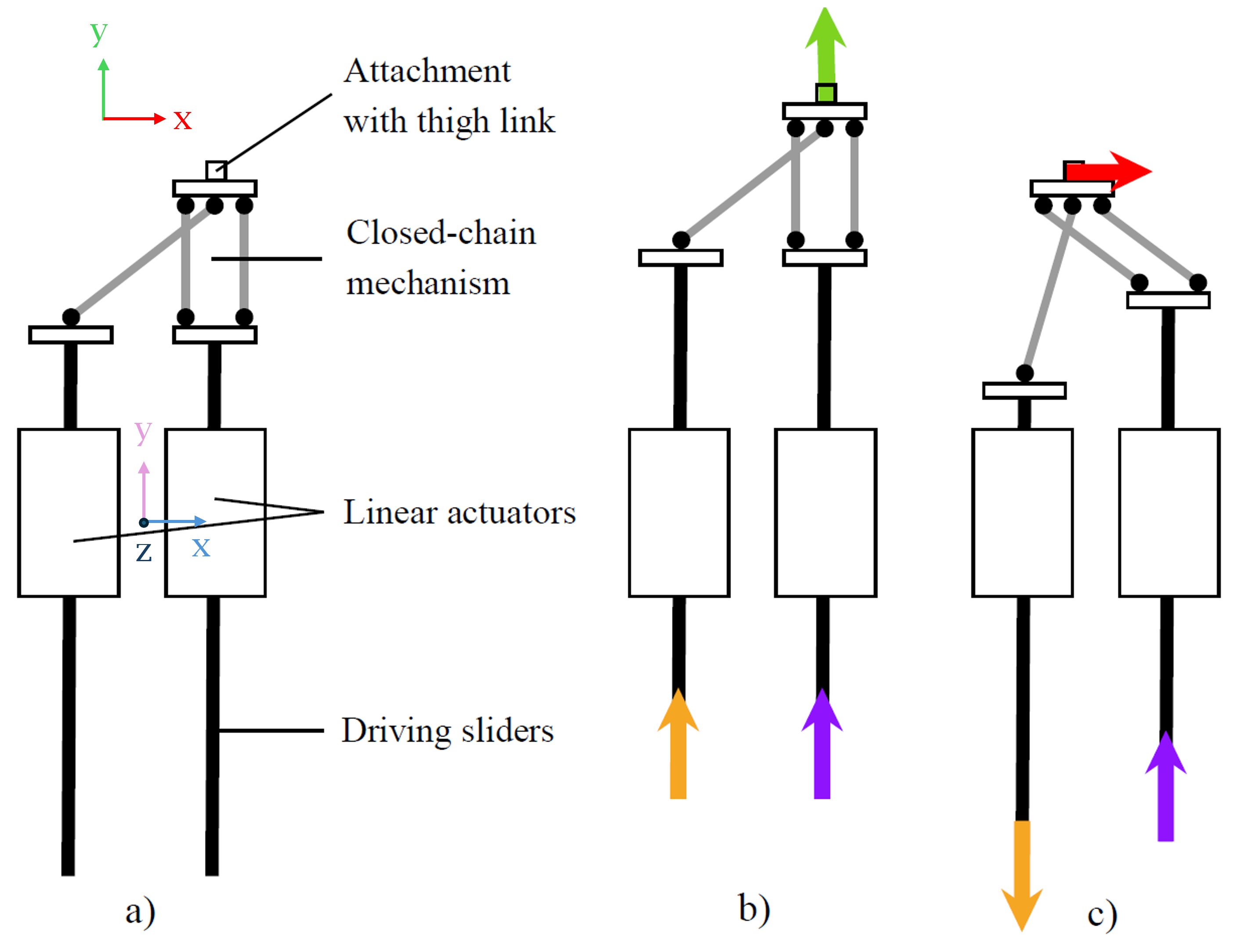}
    \caption{Top view of the right leg's closed-chain mechanism with the pelvis fixed in rotation, shown in different configurations: a) Neutral, b) Hip flexion/extension motion, and c) Hip ab-/adduction motion. Orange and purple arrows represent the movement of the linear actuators. Red and green arrows represent the resulting motion at the thigh. The blue and pink x-y-z arrows in a) represent the frame located in correspondence to the hip passive joint.}
    \label{fig:hip-model}
    \vspace{-0.3cm}
\end{figure}

\textbf{Hip actuation} is achieved through a closed-chain mechanism (Fig.~\ref{fig:experimental_setup}.c) driven by two linear actuators per leg (P01-48x240 motors, LinMot, Switzerland). 
This configuration enables control of both hip flexion/extension and ab-/adduction, while maintaining a rigid joint rotation. Its mechanism converts the independent translational motions of the linear actuators into coordinated movements of the thigh link in the frontal and sagittal planes, as illustrated in Fig.~\ref{fig:hip-model}. 
In particular, this configuration enforces hip flexion/extension movements when the two linear actuators of a leg move in the same direction (Fig.~\ref{fig:hip-model}.b), while ab-/adduction movements are produced by moving in opposite directions (Fig.~\ref{fig:hip-model}.c). By allowing the linear actuators to rotate around the x-axis of the frame located in correspondence to the hip passive joint, the system avoids overconstraining the leg. 
This mainly parallel actuation configuration offers several benefits over a serial approach% like the FreeD
, the most notable being a reduction in orthosis inertia, resulting in a more transparent system by design. % resulting  the robot a versatile system able to provide either stiff or compliant control.
%%%%%

The various actuators have built-in incremental position encoders. The pelvis module includes a Pixart infrared (IR) motion camera (PixArt Imaging Inc., Hsinchu, Taiwan) and a 6-DoF inertial measurement unit (IMU)(MPU9250, InvenSense, San Jose, USA). A Kalman filter is used to combine the IMU and camera data to estimate the position and orientation of the pelvis plate with respect to the fixed back of the pelvis module, see~\cite{wyss2019} for further details. 

The exoskeleton is attached to the participant's lower limbs using the Lokomat\textsuperscript{\textregistered} original cuffs placed on the thighs, shanks, and ankles. The participant's pelvis is secured to the pelvis module between two house-made fixtures positioned on either side, as shown in Fig.~\ref{fig:experimental_setup}.a. 
Additionally, participants wear a body harness suspended on the BWS. The knee joint is part of the original Lokomat\textsuperscript{\textregistered} device.

Since the mechanism of the exoskeleton does not incorporate any sensors to measure the rotations of the hip joint, we developed a closed-form analytical solution for the the direct and inverse mapping between the actuators' joint coordinates representing the shaft length of the two prismatic actuators of the hip module, and the space of hip joint coordinates, representing the hip flexion/extension, ab-/adduction, and internal/external rotation, respectively. The details of the mapping, together with a detailed description of the kinematic model of the exoskeleton, are presented in the \textit{Supplementary Materials}.

\subsection{Gait pattern prediction models}
\label{sec:model}

We developed models to predict leg joint and pelvis trajectories based on individual characteristics such as height and age. Our approach builds on the models proposed by Koopman \textit{et al.} \cite{koopman2014}, which use multiple polynomial regression models to predict knee and hip flexion/extension joint trajectories from a person's height and walking speed. To improve the personalization potential of Koopman's proposed models, here we also account for the effect of other factors (e.g., age, body weight, and gender), which have been shown to substantially affect joint angles and overall walking patterns~\cite{oberg1994,hanlon2006,wu2018,ren2019}. 
We further expanded the model to predict the trajectories of the augmented DoFs of our exoskeleton, namely, the lateral pelvis translation and hip ab-/adduction.

%Alongside the models to predict personalized joint trajectories, we also defined a Standard Gait Pattern, meant to providea standard pattern ideally suitable for all participants.
%Following Koopman’s approach, we used regressions to predict a sparse set of key points—called key events, namely —along each joint trajectory, rather than the full time series. We are following the same concept for our approach: for each key event, the model estimates time, joint angle or displacement, velocity, and acceleration. These events are spaced to capture the trajectory’s overall shape, allowing full personalized gait patterns to be reconstructed through interpolation.

\subsubsection*{\textbf{Database}}
\label{sec:database}
To train and validate our model, we used the publicly available gait database from the Laboratory of Biomechanics and Motor Control at the Federal University of ABC, Brazil~\cite{fukuchi2018}. This database includes data from \SI{42}{}~volunteers, split into \SI{24}{}~young adults (\SIrange{21}{37}{years}) and \SI{18}{}~older adults (\SIrange{50}{84}{years}), all without lower-extremity injuries or gait impairments. Data were collected with a marker-based motion-capture system composed of twelve cameras, from participants walking barefoot on a treadmill at eight different speeds, ranging from \SI{40}{\%} to \SI{145}{\%} of their comfortable, self-selected walking speed.
Table~\ref{tab:anthropometric_demographic_data} summarizes the anthropometric and demographic characteristics of this dataset. Compared to the database used by Koopman \textit{et al.}, which included gait kinematics from fifteen middle-aged participants, this dataset encompasses a broader range of subjects, including individuals in their twenties up to nearly \SI{80}{years} old.

\begin{table}[ht!]
\centering
\caption{Descriptive statistics of participants' anthropometric and demographic data from~\cite{fukuchi2018} used to train our predictive models}
\begin{tabular}{lllll}
\toprule
Parameter   & Mean   & Std.  & Min  & Max  \\ \midrule
Age (years) & 42.64  & 18.62 & 21   & 84   \\
Height (cm) & 167.12 & 11.01 & 147  & 192  \\
Mass (kg)   & 67.76  & 11.24 & 44.9 & 95.4 \\ \bottomrule
\end{tabular}
\label{tab:anthropometric_demographic_data}
\end{table}

The dataset includes both raw---e.g., marker coordinates and external forces---and processed data---e.g., knee and hip flexion/extension and hip ab-/adduction joint trajectories for each participant at recorded speed level. The database did not include the processed data regarding the lateral pelvis trajectories, so we derived them from the raw marker trajectories. 
All joint trajectories, including the newly computed lateral pelvis movement, were represented as time-normalized ensemble averages for each participant at their respective gait speeds.

While the database included treadmill speeds up to \SI{8.02}{\kilo\metre\per\hour}, our treadmill has a speed limit of \SI{3.2}{\kilo\metre\per\hour} \cite{lokomatPro2023}. Therefore, only a subset of the speeds recorded in the original database was included for training our gait pattern prediction model, namely:
\begin{itemize}
    \item Level 1, \SI{40}{\%} of self-selected speed: $1.80\pm$ \SI{0.231} {\kilo\metre\per\hour};
    \item Level 2, \SI{55}{\%} of self-selected speed:  $2.46\pm$\SI{0.336}{\kilo\metre\per\hour};
    \item Level 3, \SI{70}{\%} of self-selected speed: $3.14\pm$\SI{0.410}{\kilo\metre\per\hour}.
\end{itemize}

\begin{figure*}[ht!]
    \centering
    % First row: three plots side by side
    \begin{subfigure}{}
        \includegraphics[width=0.45\linewidth]{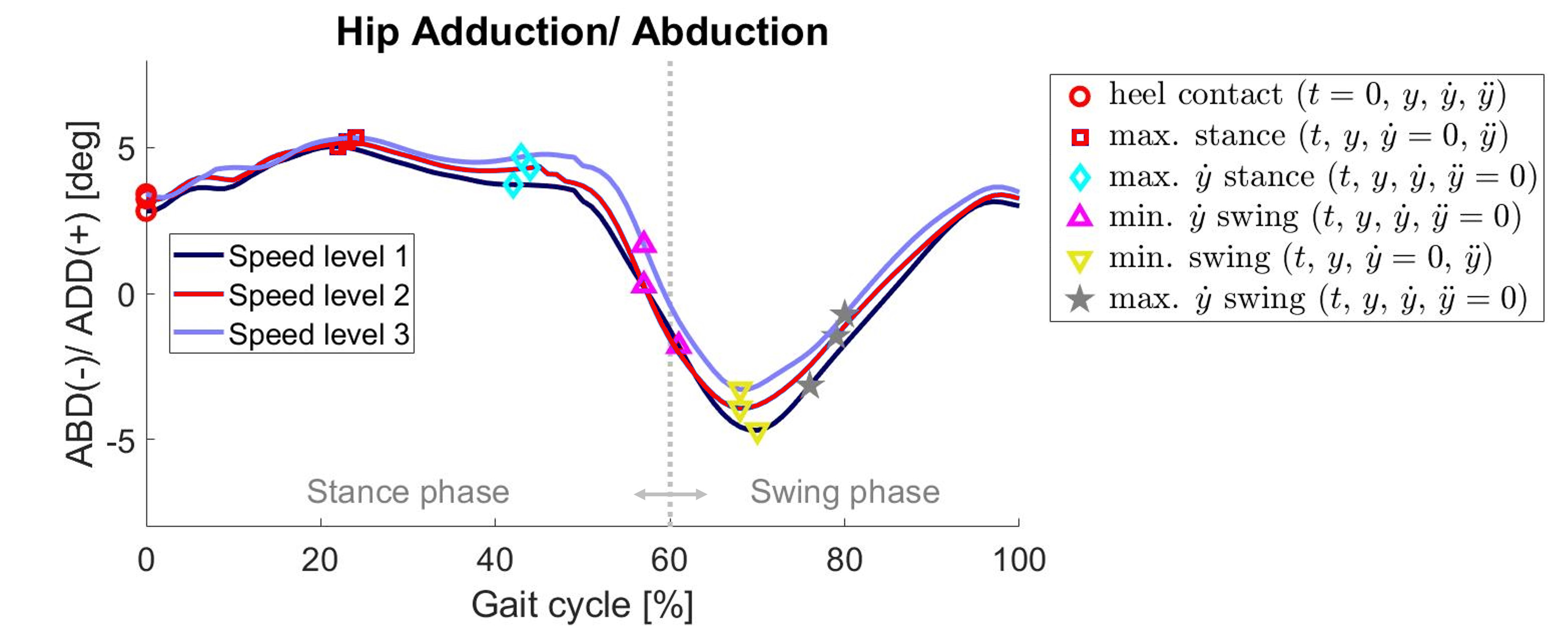}
    \end{subfigure}%
    \begin{subfigure}{}
        \includegraphics[width=0.45\linewidth]{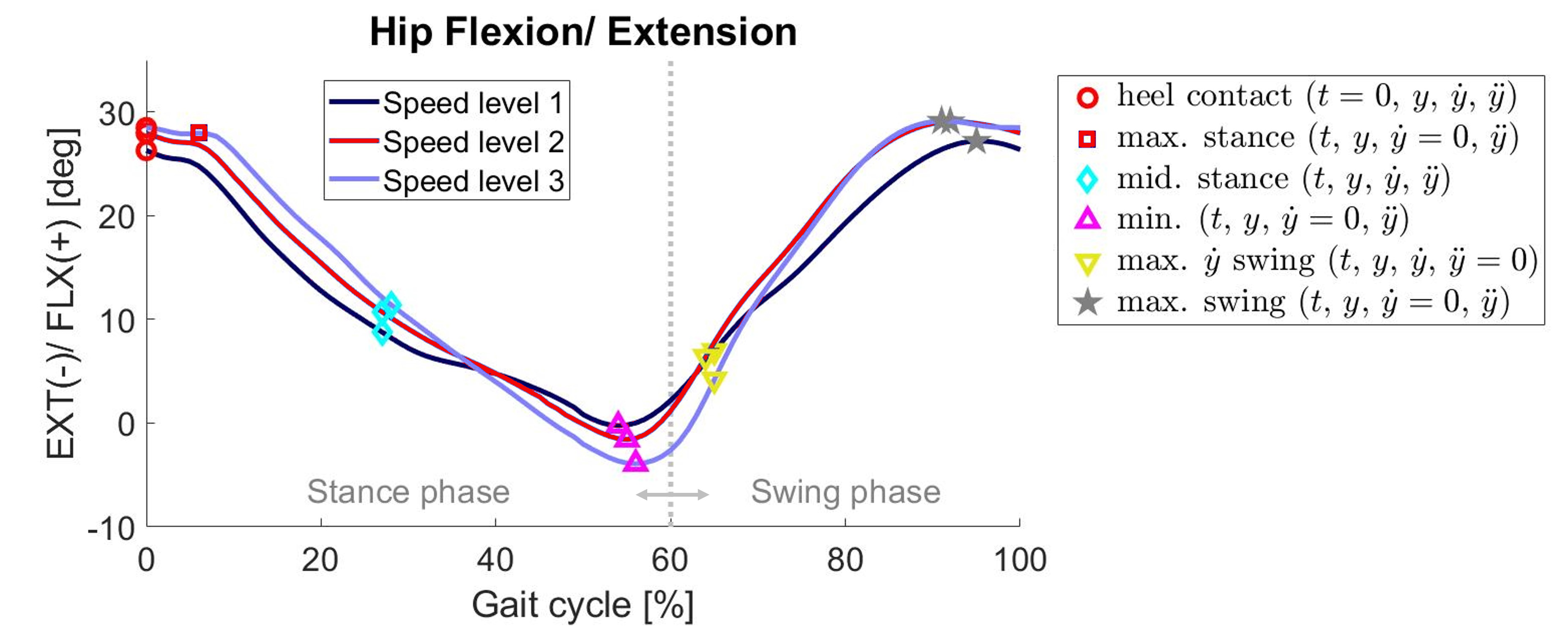}
    \end{subfigure}
    \vspace{0.5cm} \\
    
    % Second row: one plot
    \begin{subfigure}{}
        \includegraphics[width=0.45\linewidth]{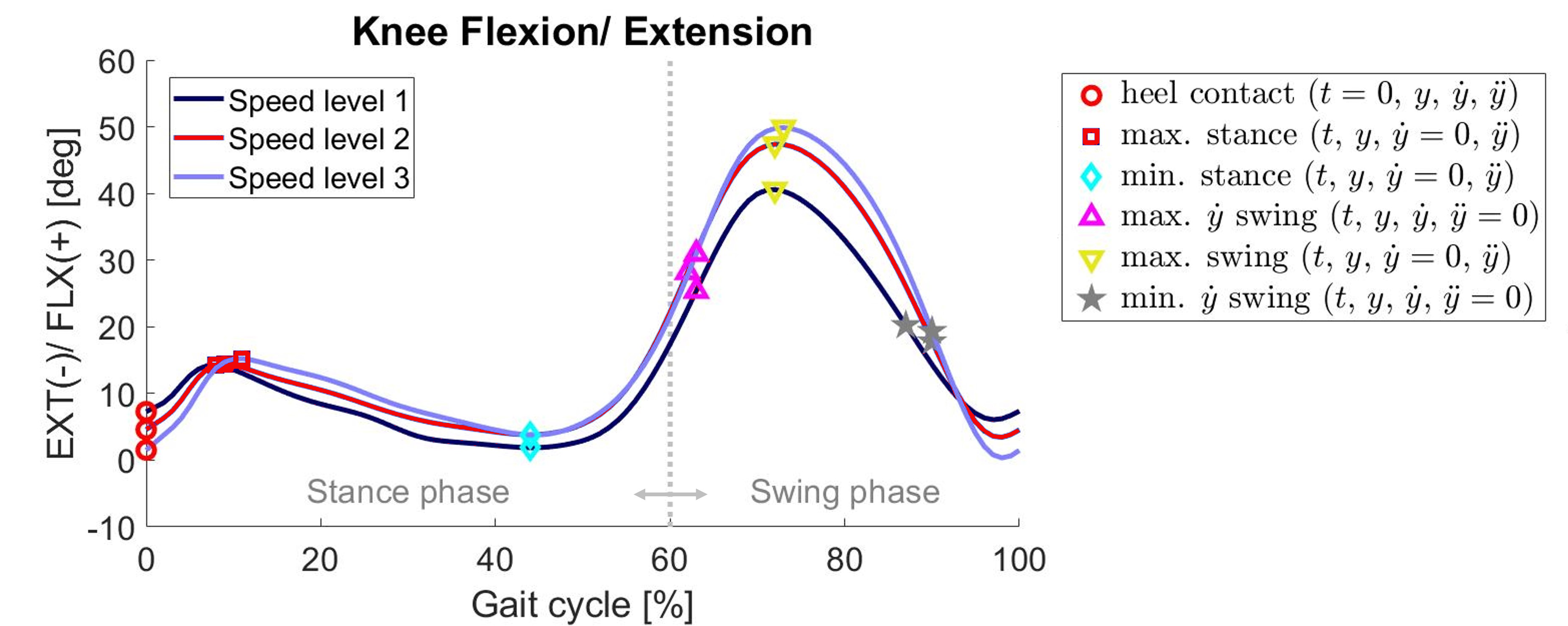}
    \end{subfigure}%
    \begin{subfigure}{}
        \includegraphics[width=0.45\linewidth]{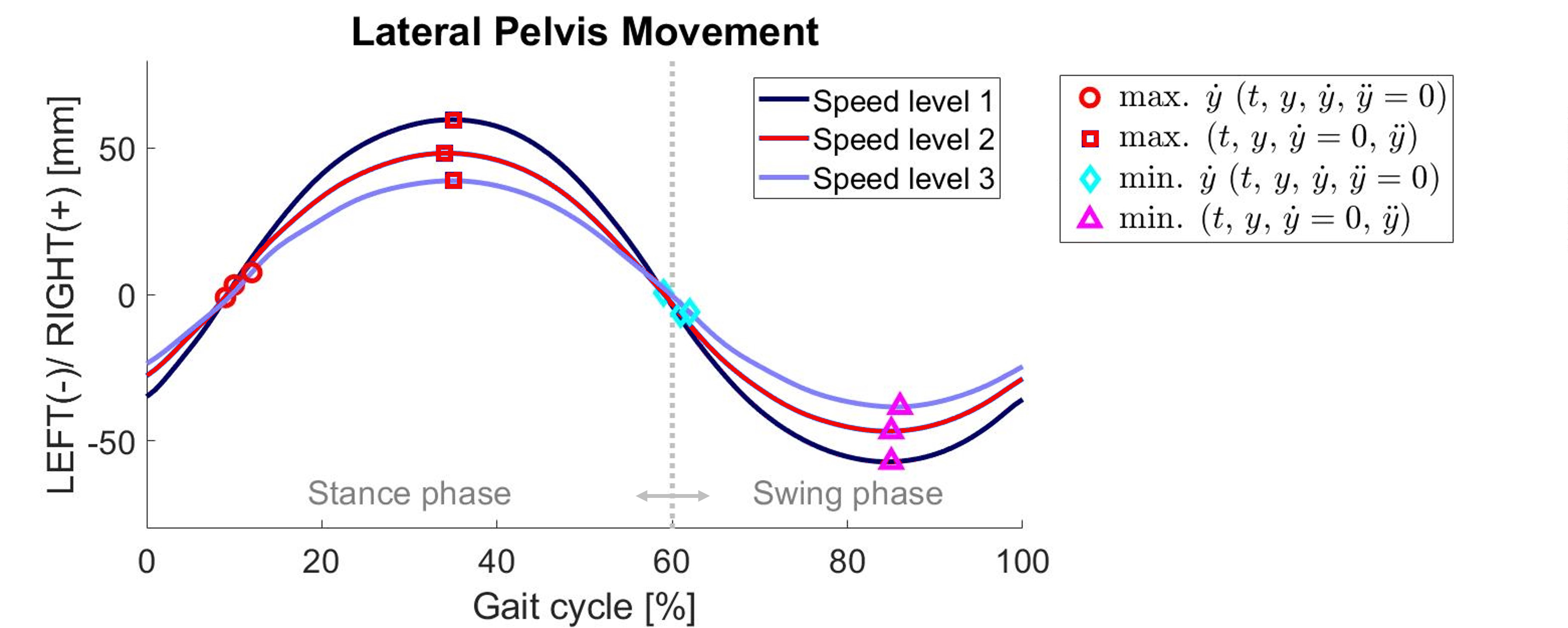}
    \end{subfigure}

    \caption{Exemplary right leg trajectories of a participant are displayed at walking speed levels 1, 2, and 3, together with their extracted key events. These key events are the same as in Koopman's study \cite{koopman2014}, except for lateral pelvis movement (not present in their study), for which extreme position and velocity values were chosen. The key events for the left and right joint trajectories were extracted separately. By default, heel strike timing is set to zero (\% of the gait cycle), as are the minimum and maximum values for joint position and velocity. These constraints are detailed in the key event legend.}
    \label{fig:key_events_example}
\end{figure*}

\subsubsection*{\textbf{Prediction models for gait key events}}
\label{sec:reg_training}

Following Koopman \textit{et al.} approach, we used regression models to predict a sparse set of key points within the joint trajectories, called key events. 
These events are spaced to capture the trajectory’s overall shape, allowing full personalized gait patterns to be reconstructed through interpolation. From the hip and knee trajectories, six key events each were selected, namely the start of the joint trajectory (heel contact) and the maximum values in position and velocity during both swing and stance phases~\cite{koopman2014}. 
For the lateral pelvis movement, only four key events were employed, corresponding to the point of maximum and minimum pelvis position and velocity along the trajectory. 
Each event is defined as a set of four parameters: (i) timing ($t$) expressed as a percentage of the gait cycle, (ii) angle or displacement ($y$), (iii) (angular) velocity ($\dot{y}$), and (iv) acceleration ($\ddot{y}$); see Fig.~\ref{fig:key_events_example} for exemplary right leg trajectories with key events of a single participant at the three walking speed levels. 

We trained regression models to predict each of these four parameters, which were then used to reconstruct the final personalized trajectories. We employed the following initial main regression equation:
\begin{equation}
\label{eq:main}
Y = \beta_c + \beta_v v + \beta_{v^2}v^2 + \beta_hh + \beta_ww + \beta_aa + \beta_ss,
\end{equation}
where $v$ represents walking speed, $h$ body height, $w$ body weight, $a$ age, and $s$ gender, encoded as a numerical value where female is $-1$ and male is $1$. The output $Y$ represents the predicted key event parameter, i.e., time ($t$), position ($\theta$), velocity ($\dot{y}$), or acceleration ($\ddot{y}$).

We fit one regression model per parameter for all the key events, leading to 24 regression models for each hip ab-/adduction, hip flexion/extension, and knee flexion/extension trajectory (i.e., four parameters per each of the six key events per trajectory), and 16 models for the lateral pelvis movement (with only four key events). To prevent overfitting and improve interpretability, stepwise regression was conducted to evaluate the significance of the predictor variables in Eq.~\ref{eq:main}. Only variables with significant effects ($p < 0.01$) were retained, following the Koopman \textit{et al.} approach~\cite{koopman2014}. Then, robust regressions with a \textit{bisquare} weighting function were employed to estimate the final regression coefficients ($\beta_x$). In this way, different models could rely on different predictors (see the \textit{Supplementary Materials} for the complete regression equations derived for the different joint trajectories).

An additional regression model was derived from Eq.~\ref{eq:main} to predict the gait cycle time, resulting in:
\begin{equation}
\label{equa:regression_equation_cycle_time}
T_{pre} =  2.7662 - 0.7458v + 0.0903v^2  - 0.0037a, 
\end{equation}
where the gait cycle time ($T_{pre}$) depends on the walking speed ($v$) and age ($a$).

\subsubsection*{\textbf{Generation and evaluation of Personalized Gait Patterns based on predicted key events}}

From the predicted four parameters of each gait key event (i.e., $t$, $y$, $\dot{y}$, or $\ddot{y}$), we then generated personalized continuous joint kinematic trajectories. We employed the 5th-order piece-wise quintic splines interpolation method, as proposed by Koopman \textit{et al.}~\cite{koopman2014}.

The accuracy of the predicted trajectories was evaluated using the data from the publicly available database \cite{fukuchi2018}. We computed the root mean square error (RMSE) against the real database trajectories, using the leave-one-out cross-validation method. The results were then averaged across all participants and over both left and right joint trajectories.

% The RMSE is defined as follows:
% \begin{equation}
% RMSE = \sqrt{\frac{\sum_{t=1}^{N} (y_t - \hat{y}_t)^2}{N}}.
% \label{equation:RMSE}
% \end{equation}
% Here, $N$ represents the number of data points in the gait cycle (101 time-normalised points), $y_t$ corresponds to the actual joint angle/displacement value at time $t$, and $\hat{y}_t$ denotes the corresponding predicted value. %This evaluation method allows for a quantitative assessment of the performance of the gait prediction model in terms of accuracy in predicting gait patterns.

\subsubsection*{\textbf{Generation and evaluation of the \textit{Standard Gait Pattern}}}

Alongside the \textit{Personalized Gait Pattern}, we used the same database to define a \textit{Standard Gait Pattern}, meant to provide a standard pattern ideally suitable for all participants. This standard pattern was obtained by averaging the gait patterns of the three included walking speed levels across all participants in the dataset. We globally assumed symmetrical gaits, so the trajectories of both the left and right legs were combined. Note that this average was taken over the entire trajectory, rather than averaging over the key events. 

Because this was conceived as a standard gait pattern for all scenarios, we, therefore, had to adjust the trajectories by estimating the gait cycle time ($T_{stand}$) to account for different desired gait speeds. Therefore, a regression model, reduced following the same steps described above, was employed to estimate the gait cycle time also for the \textit{Standard} gait pattern: 
\begin{equation}
\label{equa:regression_formula_cycle_time_standard}
\begin{aligned}
T_{stand} = \ & 1.8993 - 0.6909v + 0.0789v^2 + 0.3928h. %
\end{aligned}
\end{equation}

We calculated the RMSEs between the standard and actual gait patterns from the database 
as a means to compare their accuracy with that from the \textit{Personalized} gait pattern.

\subsection{Pilot user evaluation of the Personalized and Standard gait patterns}
\label{sec:experiment}
\subsubsection*{\textbf{Participants}}
Ten unimpaired participants participated in the pilot study, equally divided into females and males. Their ages ranged from \SIrange{23}{27}{years}, with a mean age of $25$ years, average height of $1.76\pm$\SI{0.089}{\meter}, and weight of $69.25\pm$\SI{12.79}{\kilo\gram}. None reported neurological or orthopaedic disorders. 
%
%
%\subsubsection*{\textbf{Experimental protocol}}
The study was approved by the TU Delft Ethics Committee (HREC, Application 25-08-2023) and was conducted in accordance
with the Declaration of Helsinki. Prior to the experiment, all participants read and signed an informed consent form. % (application date: 25.08-2023). 

\subsubsection*{\textbf{Experimental protocol}}
Participants underwent three different test conditions (trials), of two minutes each, with the exoskeleton guiding their lower limbs along three distinct gait patterns: (i) the \textit{Personalized} gait pattern, (ii) the \textit{Standard} gait pattern, and (iii) a \textit{Random} gait pattern. The order of the gait pattern condition trials was pseudo-randomized across participants, i.e., each of the six possible condition sequences was used once (following a Latin square), and then four more sequences were randomly selected from these six. 

The first two gait patterns were introduced in detail in Sec.\ref{sec:reg_training}. For the \textit{Random} gait pattern condition, each participant was assigned a gait pattern randomly selected from the recorded patterns from Fukuchi's database \cite{fukuchi2018}. The sampling was restricted to patterns that fell within the previously selected subset of speeds.
The inclusion of this random pattern
served as a baseline to observe the effects of non‑tailored but still physiologically plausible assistance, resembling current clinical practice where pre-recorded gait patterns are used as input to trajectory-based controllers.% By using realistic human gait patterns that (might) differ from each participant’s own kinematics, the \textit{Random} condition was designed to introduce a controlled mismatch, sufficiently different from the participant's natural gait, yet not so physiologically distant to produce extreme discomfort.
%The order of the gait pattern condition trials was pseudo-randomized across participants, i.e., each of the six possible condition sequences was used once (following a Latin square), and then four more sequences were randomly selected from these six.

Participants were instructed to remain passive to the movement of the exoskeleton and received no indication about the differences among gait pattern conditions. For all conditions, the walking speed was set to \SI{1.8}{\kilo\metre\per\hour}, reflecting the average speed of the lowest walking speed level in the database. %Therefore, the random gait pattern was selected only from the pool of patterns in the lowest walking speed level, i.e., Level 1 in Sec.\ref{sec:database}. 
Only for some participants, who explicitly requested to increase the speed to walk more comfortably, it was set to  \SI{2.0}{\kilo\metre\per\hour}. While participants were wearing the harness for safety, the BWS system remained slack, such that participants did not receive any weight compensation.

We operated the exoskeleton in Proportional-Derivative (PD) control, with the computed gait patterns as reference trajectories. The controller was intentionally rigid (for the hip and pelvis PD controller: $Kp =$\,\SI{5.0}{\ampere\per\milli\meter} and $Kd =$\,\SI{7.5}{\ampere\per(\meter\per\second)}, for the knee PD controller: $Kp =$\,\SI{1.0}{\ampere\per\milli\meter} and $Kd =$\,\SI{0.1}{\ampere\per(\meter\per\second)}), allowing minimal compliance to enhance the user's sensitivity to variations in gait patterns. We also aimed to achieve stiff connection points between the participants' limbs and the robotic device at the pelvis, thighs, and shanks to make trajectory differences more perceptible to the participants. %Yet, we acknowledge that the soft tissue of participants' limbs, together with the Velcro\textsuperscript{\textregistered} straps used at the human-robot interaction points, might have introduced some compliance in the connection. 

%For all conditions, the walking speed was set to \SI{1.8}{\kilo\metre\per\hour}, reflecting the average speed of the lowest walking speed level in the database. %Therefore, the random gait pattern was selected only from the pool of patterns in the lowest walking speed level, i.e., Level 1 in Sec.\ref{sec:database}. 
%Only for some participants, who explicitly requested to increase the speed to walk more comfortably, it was set to  \SI{2.0}{\kilo\metre\per\hour}. % The walking speed for the random patterns varied around the target speed of \SI{1.8}{\kilo\metre\per\hour}, with a standard deviation of $\pm$\SI{0.231}{\kilo\metre\per\hour}. 

After signing the informed consent form, participants wore the exoskeleton, which was adjusted to each of their body dimensions, including the width of the pelvis, the frontal position of the hip joint, thigh length, and shank length. The kinematic model of the exoskeleton was adjusted according to these dimensions. The reference actuator position trajectories were then generated using the exoskeleton model (see \textit{Supplementary Materials}) and based on hip ab-/adduction, hip flexion/extension, and pelvis lateral motion calculated patterns. Once the exoskeleton was fit and the exoskeleton kinematic model adjusted, participants underwent a familiarization phase lasting around two minutes before testing the three different gait pattern conditions. During this initial phase, participants walked with the exoskeleton in a `transparent mode'~\cite{marchalCrespo2009}, i.e., they could move as freely as possible while getting used to walking with the exoskeleton. %To ensure safety, participants wore a harness, but no body weight support was provided. 
Emergency stops were provided to both the participant and the experimenter.

\subsubsection*{\textbf{Data acquisition and processing}}

%\subsubsection{User experience}
After each condition, participants filled in a questionnaire to evaluate their perceptions of the different gait pattern conditions with regards to:
\begin{itemize}
    \item \textbf{Interest/Enjoyment:} Measured with four items from the Intrinsic Motivation Inventory (IMI) \cite{IMIhompage2023}, with scores averaged per IMI guidelines.
    \item \textbf{Comfort:} Assessed through seven self-designed questions to capture various aspects of comfort experienced during exoskeleton use.
    \item \textbf{Naturalness:} Evaluated with four self-designed questions about how participants perceived the movements of the exoskeleton, analyzed individually.
    \item \textbf{Passiveness:} Evaluated through four self-designed items to determine whether participants remained passive using the exoskeleton, with scores averaged to assess overall passiveness.
\end{itemize}

The full list of questions/items employed can be found in the \textit{Supplementary Materials - Table VI}. Participants filled in the questionnaires following a 7-point Likert scale, where, according to the question: (1) Very unnatural/uncomfortable/dissimilar - (7) Very natural/comfortable/similar or (1) Not true at all - (7) Very true. 

At the end of the experiment, participants also ranked---from (1) Favorite to (3) Least favorite---the three gait patterns in terms of overall experience, comfort, and naturalness. They also rated their confidence in these rankings, from (1) Not confident to (10) Very confident. Finally, we included open-ended questions (reported in the \textit{Supplementary Materials}) to capture additional insights and comments.

%\subsubsection{Sensors data}
Besides the participants' subjective experience, to determine whether walking along certain gait patterns led to increased human-robot interaction forces, the force measured by the original knee joint force sensor was recorded at a sampling frequency of \SI{100}{Hz} during the experiment. The \textbf{Mean Absolute Force} was then calculated per condition and leg, defined as the average magnitude of the force exerted at the right/left knee joints.

\subsubsection*{\textbf{Data analysis}}
\label{Sec:analysis}
    
We evaluated the effect of the different gait pattern conditions on the subjective and objective performance variables using a linear mixed model (LMM) of the form:
\begin{equation}
\label{eq:interaction}
Var = Pattern + Trial Order + (1|Participant),
\end{equation}
where $Var$ refers to any of the measured performance metrics (e.g., Comfort). The independent variable $Pattern$ refers to the pattern condition (i.e., \textit{Personalized}, \textit{Standard}, or \textit{Random}). The \textit{Standard} condition was chosen as the base condition, being equivalent for all the participants. To control for potential effects related to the sequence in which conditions were presented, we also included the $TrialOrder$ as an independent variable in the model. This ordinal variable represents the position of each pattern condition within the experimental sequence for each participant (i.e., \textit{First}, \textit{Second}, or \textit{Third}). We also included \textit{Participant} as a random factor. We fit the model using the \texttt{lmer} function from the \texttt{lme4} package in \texttt{R}.

Additionally, ranking outcomes were analyzed using a Friedman test, comparing across experimental conditions and trial order. After identifying significant differences with the Friedman test, a post-hoc Nemenyi test was conducted to explore pairwise comparisons between conditions. This analysis was performed using the \texttt{frdAllPairsNemenyiTest} function from the \texttt{PMCMRplus} package in \texttt{R}. We set all significance levels to $p<0.05$ and corrected for multiple comparisons with Bonferroni correction.

\section{Results}
\subsection{Accuracy evaluation of the Personalized and Standard gait patterns}

Results from the leave-one-out cross-validation method used to assess the RMSEs between the \textit{Personalized}/\textit{Standard} and actual gait patterns are reported in Table~\ref{tab:rmse_gait_prediction_model}. 
We did not observe large differences in accuracy between the  \textit{Personalized} and \textit{Standard} gait patterns. For the hip joint trajectories, the \textit{Standard} pattern's RMSE values were slightly lower compared to the \textit{Personalized} ones. Conversely, for the knee and pelvis joint trajectories, the \textit{Personalized} reconstructed trajectories had slightly lower RMSE than the \textit{Standard} trajectories.

\begin{table}[ht!]
\caption{RMSE between the \textit{Personalized}/\textit{Standard} joint trajectories and actual (Act.) measured trajectories.}
\begin{tabularx}{\columnwidth}{p{2.8cm}XX}
\toprule
Joint & 
\begin{tabular}[c]{@{}l@{}}RMSE%\textsuperscript{a}
\\ Act-Personalized\end{tabular} & \begin{tabular}[c]{@{}l@{}}RMSE%\textsuperscript{a}
\\ Act-Standard\end{tabular}\\ \midrule
Hip abd/add (deg) & 2.906 & 2.772 \\
Hip flex/ext (deg) & 7.573  & 6.953 \\
Knee flex/ext (deg) & 5.809  & 6.385 \\
Pelvis lateral (mm) & 6.321 & 6.779 \\
\bottomrule
\end{tabularx}
%\vspace{-0.5cm}
\label{tab:rmse_gait_prediction_model}
\end{table}

\subsection{User experience}
Results from the LMM for the user experience analysis, %are reported in Table~\ref{tab:lmmResultsAll}. The 
together with the average scores assigned per gait pattern condition are reported in the \textit{Supplementary Materials - Table VII and Figure 4}. Here, we comment on relevant results.

In terms of \textit{Comfort}, we found that the overall comfort significantly improved in the third trial with respect to the first one ($\beta = 1.43,\ t = 2.94,\ p = 0.01$). No significant differences were perceived among conditions. When looking into the individual \textit{Comfort} sub-questions, we found a significant effect of the gait pattern condition on the perceived \textit{Physical strain}. In particular, participants reported higher \textit{Physical strain} when walking with the \textit{Personalized} gait pattern vs. the \textit{Standard} pattern condition ($\beta = 1.07,\ t = 2.86,\ p = 0.011$). Interestingly, \textit{Physical strain} seemed to decrease with trials, albeit non-significantly ($\beta = -0.26,\ t = -0.70,\ p >0.05$).

Regarding \textit{Naturalness}, we also found a main effect of the condition order. In particular, the third trial was perceived significantly as more natural than the first trial ($\beta = 1.35,\ t = 2.79,\ p = 0.013$), and as more similar to the participants' own way of walking ($\beta = 1.85,\ t = 3.59,\ p = 0.002$). Questions related to the \textit{Smoothness of movements} and whether the limbs were pushed beyond their natural range did not show significant differences across conditions or condition order.

The models for both the \textit{Interest/Enjoyment} (IMI) and \textit{Passiveness} metrics revealed no significant differences, either in terms of gait pattern or condition order. The \textit{Personalized} pattern got the highest enjoyment score, but differences with the other patterns were not significant ($\beta = 0.20,\ t = 1.35,\ p > 0.05$).

When analyzing self-reported condition rankings, we did not find significant differences across the gait patterns, including overall preference, comfort, and naturalness (see Table~\ref{tab:friedmanTestResults}). However, we found that the order of the trials seemed to play a role in the participants' rankings. In particular, we found that the third trial was generally perceived as more comfortable ($p = 0.037$) and more natural ($p = 0.037$) compared to the first trial. This comes with a general high confidence in participants' rankings, scoring $8.2 \pm 0.92$ regarding overall preference, $8.3 \pm 1.06$ regarding the most comfortable gait pattern, and $7.6 \pm 1.51$ regarding the most natural gait pattern, on a scale from $1$ (not confident at all) to $10$ (very confident). For completeness, the responses to the open-ended questions are included in the \textit{Supplementary Materials}.

\begin{table*}[hb!]
\caption{Friedman test results for participants' rankings. When the test was significant, we performed pairwise comparisons across patterns - Standard (S), Personalized (P), and Random (R) - or across experimental trials -- 1st: T1, 2nd: T2, 3rd: T3. *($p < 0.05$)}
\label{tab:friedmanTestResults}
\begin{tabularx}{\textwidth}{p{3.5cm} p{3.5cm} *{6}{>{\centering\arraybackslash}X}}
\toprule
\multirow{2}{*}{Comparison Groups} & \multirow{2}{*}{Ranking}  
& \multicolumn{3}{c}{Friedman test} 
& \multicolumn{3}{c}{Post-hoc comparison $p$-values} \\ 
\cmidrule(lr){3-5} \cmidrule(lr){6-8}
& & $\chi^2$ & df & $p$-value & S vs. P & S vs. R & P vs. R \\ 
\midrule
\multirow{3}{*}{\shortstack{Conditions \\ (Patterns)}} 
& Overall preferred pattern & 4.1 & 2 & 0.122 & - & - & - \\
& Most comfortable pattern  & 3.2 & 2 & 0.202 & - & - & - \\
& Most natural pattern      & 2.6 & 2 & 0.272 & - & - & - \\ 
\cmidrule(lr){3-5}\cmidrule(lr){6-8}
& & $\chi^2$ & df & $p$-value & T1 vs. T2 & T2 vs. T3 & T1 vs. T3 \\ 
\midrule
Trial order 
& Overall preferred pattern & 5.6 & 2 & 0.061 & - & - & - \\ 
& Most comfortable pattern  & 6.2 & 2 & \textbf{0.045}* & 0.261 & 0.644 & \textbf{0.037}* \\ 
& Most natural pattern      & 6.2 & 2 & \textbf{0.040}* & 0.261 & 0.644 & \textbf{0.037}* \\ 
\bottomrule
\end{tabularx}
\end{table*}

\subsection{Human-robot interaction forces}
The results for the LMM analysis of the mean absolute force at both knee actuators are reported in Table~\ref{tab:lmmResultsForce}.
We found a significantly higher force when participants were enforced to follow the \textit{Random} gait pattern when compared to the \textit{Standard} gait patterns (Right leg: $\beta = 68.7,\ t = 3.6,\ p = 0.002$; Left leg: $\beta = 84.5,\ t = 4.1,\ p = 0.001$).  The difference between the \textit{Personalized} and \textit{Standard} gait patterns only approached significance on the left leg (Right leg: $\beta = 22.2,\ t = 1.2,\ p = 0.257$; Left leg: $\beta = 42.1,\ t = 2.0,\ p = 0.061$), with higher forces for the \textit{Personalized} trajectories. We did not find a significant effect of the trial order on the interaction forces.
\begin{table*}[b]
\caption{Linear mixed model results for mean absolute force at the knee actuators}
\label{tab:lmmResultsForce}
\begin{tabularx}{\textwidth}{X *{8}{l}}
% First question row
\toprule
\textbf{\multirow{2}{*}{Variable}} & \multicolumn{4}{l}{Right Knee: Mean Absolute Force (N)} & \multicolumn{4}{l}{Left Knee: Mean Absolute Force (N)} \\ 
\cmidrule(lr){2-5} \cmidrule(lr){6-9}
& \textbf{Estimate} ($\beta$) & \textbf{Std. Error} & \textbf{t value} & \textbf{p-value} & \textbf{Estimate} ($\beta$) & \textbf{Std. Error} & \textbf{t value} & \textbf{\textit{p}-value} \\ 
\cmidrule(lr){1-9}
(Intercept) & 153.4 & 20.5 & 7.5 & \textbf{8.22E-08}*** & 140.0 & 24.1 & 5.8 & \textbf{5.78E-06}*** \\
Personalized pattern & 22.2 & 18.9 & 1.2 & 0.257 & 42.1 & 20.9 & 2.0 & 0.061 \\
Random pattern & 68.7 & 18.9 & 3.6 & \textbf{0.002}** & 84.5 & 20.9 & 4.1 & \textbf{0.001}** \\
2nd Trial & -19.7 & 18.9 & -1.0 & 0.313 & -17.7 & 20.9 & -0.8 & 0.409 \\
3rd Trial & -25.6 & 18.9 & -1.4 & 0.193 & -19.2 & 20.9 & -0.9 & 0.372 \\
\bottomrule
\end{tabularx}

\vspace{0.15cm} 
\hspace{0.15cm}
\raggedright **($p <0.01$), ***($p <0.001$) 
\vspace{-0.5cm}
\end{table*}

%From the evaluation of the mean absolute error (MAE) across the various gait patterns and actuators, we found some main effects of the pattern condition and condition order (see summary of results in Table XX in \textit{Supplementary Materials}). Nevertheless, no unique trend could be detected. The largest model-estimated MAE was 8.48 mm for the left outer prismatic actuator and 3.06\degree for the left knee revolute actuator, under the standard condition in the first trial. 

\section{Discussion}

\subsection{A novel exoskeleton for enhanced degrees of freedom}

We presented a novel exoskeleton and accompanying
validated kinematic model based on a modified Lokomat\textsuperscript{\textregistered}. The device includes the standard configuration with augmented hip ab-/adduction and full pelvis translation and rotation (with actuated lateral translation), allowing for more realistic gait training. We also implemented a gait personalization algorithm, based on a comprehensive walking database, that can predict individualized gait patterns based on individuals' walking speed, anthropometric, and demographic data. 
We evaluated the potential of our system's personalized trajectories against non-personalized ones in a pilot user experience study with ten unimpaired participants.

%%OVERALL EXPERIENCE
The results of the overall user experience indicate that both comfort and naturalness could be improved, regardless of the enforced gait pattern. 
Participants' feedback indicated that there was significant discomfort associated with the device, particularly at the ankle cuff. Moreover, participants commonly mentioned they were experiencing excessive and unnatural lateral pelvis movement. This could be due to the compliance of the pelvis module itself. Instructing participants to remain passive---that they seemed to achieve based on the results from the \textit{Passiveness} questionnaire---may have caused them to let the exoskeleton sway their upper bodies side to side, potentially increasing the sensation of excessive lateral movement. 
The relatively slow walking speed enforced by the exoskeleton might have also contributed to the general unnatural feeling, as the average comfortable speed in healthy adults is around \SI{5.0}{\kilo\metre\per\hour} \cite{BAROUDI202269}. 
Nevertheless, as discussed in more detail below, the overall experience improved significantly over time, and therefore, results regarding overall experience with the exoskeleton within this pilot study should be taken with care.

\subsection{Personalizing gait kinematics may have a minimal effect on user experience} 

Contrary to our expectations, we did not find notable differences in the reported user experience---namely \textit{Interest/Enjoyment}, overall \textit{Comfort}, and \textit{Naturalness}---between the \textit{Personalized}, \textit{Standard}, and \textit{Random} gait patterns. We only found a small, albeit significant, higher perceived physical strain in the \textit{Personalized} vs. \textit{Standard} pattern. 
This increased physical strain could be due to the fact that the prediction model predicts the different joints' trajectories separately, without taking into consideration their inter-joint coupling during walking. This might lead to a reduction in enforced inter-joint coordination. Nevertheless, this perceived physical strain is not reflected in significant differences between these two specific conditions in the human-robot interaction forces, nor did participants report a lack of coordination during the open questions interview. Only a non-significant increased force at the left knee was observed in the \textit{Personalized} vs. \textit{Standard} gait pattern, while differences did reach significance in both legs between the \textit{Random} and \textit{Standard} patterns, with larger forces associated with the former. Yet, these interaction forces were only measured at the knee joint, and therefore, do not capture the whole physical strain arising from potentially suboptimal inter-joint coupling.

While healthy gait is, overall, unique and repeatable, the natural fluctuations in human motor control could explain why preference was similar across the three profiles. 
Healthy gait exhibits non-negligible stride-to-stride variability~\cite{Moreira2021, HAUSDORFF2007555} and is robust to changes in speed, environment, and perturbations~\cite{Rábago2015}. Kinematic variability also increases at slower than preferred walking speeds, such as the speeds experienced in our setup~\cite{Terrier2003,Huiying2009}. This implies that one person's average gait trajectory is not rigidly distinct from another's but is instead likely to fall within the broader distribution of kinematic patterns expressed by other healthy young adults. Such overlap reflects the redundancy and flexibility of the neuromotor system, which allows different movement strategies to achieve stable locomotion~\cite{STERGIOU2011869}. Thus, the result that participant preference did not vary between the personalized and non-personalized conditions may reflect this balance between individuality and robustness. 
%
%We may expect different results in impaired populations, in which gait variability is elevated relative to healthy young adults~\cite{MOON2016197}. Automatic personalization of therapy---such as through human-in-the-loop robotic tuning~\cite{Slade2024}---is important to address the diverse and evolving needs of patients throughout rehabilitation.

Additionally, systemic differences between the imposed and real-world gaits may overwhelm differences between the three conditions. Exoskeleton locomotion differs qualitatively from normal, unconstrained gait~\cite{Swank2019}, so users may not consciously perceive the nuances between patterns. For example, people without prior experience with exoskeletons prefer lower levels of assistance~\cite{Swank2019, Ingraham2022}, suggesting that users may be more affected by the intrusiveness of the rigid exoskeleton guidance. %The trajectories were also separately determined for each degree of freedom. This decoupling of joint kinematics may degrade the inter-joint coordination necessary for stable gait~\cite{chiu2013variability}, which could explain the lower scores, especially for the personalized pattern. 
%We, therefore, cannot decouple the effects of the pattern definition from any possible benefits of personalization.

\subsection{User adaptation to the experimental setup seems to have a stronger effect on user experience than personalization}

It is important to note that, while we did not capture significant differences in the user experience between gait pattern conditions, we still found differences between trials. The user experience seemed to improve in terms of overall comfort, naturalness, and smoothness of the exoskeleton's movements as the experiment advanced, reaching significant differences between the last and first conditions, regardless of the gait pattern enforced. These differences suggest that participants probably adapted to the system over time, leading to an enhanced walking experience. This is supported by the overall higher rating of the last trial as the most comfortable and natural pattern compared to the first one, while no differences in rating were found between gait patterns. 

These ratings raise questions about whether participants actually felt differences between gait patterns, despite their overall confidence in their responses. Indeed, when looking at the accuracy of the \textit{Personalized} and \textit{Standard} patterns with respect to the actual patterns from the starting database (see Table~\ref{tab:rmse_gait_prediction_model}), they both showed similar RMSE for the joint trajectories. This, together with the compliance of the human-robot interfaces, namely the Velcro\textsuperscript{\textregistered} straps and soft leg tissue, might have muted the perception of the different gait patterns. 

Exposure is known to be an important factor underlying human-robot performance~\cite{Ingraham2022, Poggensee2021, Tankink2025}. While some studies indicate that significantly more time is required to adapt to exoskeleton assistance~\cite{GORDON20072636, Poggensee2021}, these results suggest that people begin to enjoy the device more within only a few minutes. The paradigm presented in this study importantly differs from those other exoskeletons---as the goal was to fully guide the user rather than to partially augment existing capabilities---although our results still highlight the need to consider user adaptation when designing human-robot interaction protocols. Adaptation occurs over several different timescales, with kinematic adaptation occurring fairly quickly~\cite{GORDON20072636}. It was for this reason that we included a first familiarization period of around two minutes. We then chose two-minute trials for each condition, balancing the time required to form an opinion of the pattern at hand while reducing the potential for fatigue that could occur had the experiment been too long. %Nevertheless, this pilot study could have benefited from a longer familiarization period to account for this adaptation in preference and comfort.

\subsection{Study limitations}

Our work presents limitations related to the employed prediction model and the design of the pilot study. First, the dataset utilized in this study, from Fukuchi~\cite{fukuchi2018}, %primarily comprises data from the Brazilian population,
may not fully represent the anthropometric variability observed in broader populations (e.g., differences in height~\cite{datapandas2024}), which could affect the generalizability of the results.
Second, although two-minute trials represented a reasonable compromise, a longer familiarization period may have better accounted for possible adaptation to the experimental setup. This should be accompanied by enough resting time between familiarization and the start of the first condition. 
% 
%The study included only unimpaired participants, limiting applicability to individuals with gait impairments, such as those following acquired brain injury. 
The small sample size, the treatment of ordinal Likert data as continuous, and the presence of an order effect might also limit the robustness of the statistical analysis.

Participants’ feedback suggests that the reference gait patterns were perceived as unnatural and uncomfortable. Such perceptions are likely influenced not only by the patterns themselves but also by the mechanical structure and rigid control strategy of the robotic system. Factors that may have masked personalization effects include the specific cuff design, the intrusiveness of full trajectory guidance, and the independent prediction of joint trajectories. The soft tissue of participants' limbs, together with the Velcro\textsuperscript{\textregistered} straps used at the human-robot interaction points, might have introduced some compliance in the connection and altered the perception of specific patterns. Uncontrolled factors, such as personal footwear---standardized in the reference dataset~\cite{fukuchi2018}---may also have influenced perception. Importantly, the rigid position-derivative controller, selected intentionally to amplify the perceptibility of the trajectories, might itself have contributed to participants' discomfort. While rigid guidance was meant to emphasize differences, it also reduced compliance at the joints. Overall, this might have determined a general sensation of unnaturalness that overshadowed the subtler distinction between patterns. 

Such restrictive behavior is a general drawback of trajectory‑based exoskeleton control~\cite{Zhang_energy}, especially when interacting with unimpaired users~\cite{de2023control}. To overcome this limitation, emerging trajectory‑free frameworks offer an alternative perspective on how personalization can be defined and evaluated. For example, energy-shaping controllers~\cite{Zhang_energy,slade_personalizing_2022} personalize the assistance by modifying dynamics that are not task-dependent, softening the reliance on predefined joint trajectories. %In the realm of neurorehabilitation, neuromuscular control strategies use biosignal recordings, such as muscular—electromyography (EMG) (e.g.,~\cite{XX}) or brain—electroencephalography (EEG) signals (e.g.~\cite{xx}) to handle the control objective.
%The combination of trajectory-tracking control or neuromuscular control with more compliant control usually provides a more flexible behavior to the exoskeleton during rehabilitation e.g., by allowing more movement variability around the desired trajectory, compared to conventional rigid Low-level controllers such as proportional-derivative (PD) controllers~\cite{de2023control}. 
Additionally, more compliant control strategies can mitigate the overall discomfort observed with rigid robotic guidance by allowing natural variability around the desired trajectory~\cite{de2023control}. However, these approaches typically require a degree of shared autonomy with the user, which may not be suitable in high-risk environments or for individuals with limited physical capacity. For example, neuromuscular interfaces based on brain electroencephalography (EEG; e.g.~\cite{benabid_exoskeleton_2019}) or muscular electromyography (EMG; e.g.,~\cite{Tan2018}), when combined with compliant control, can further enhance adaptability, but also increase reliance on user-generated inputs~\cite{abbink_haptic_2012}. Nevertheless, in the specific case of exoskeletons to assist/rehabilitate patients with severe sensorimotor impairments, the users might find rigid guidance ``reassuring" or ``stable," and therefore, benefit more from gait pattern personalization. These approaches further demonstrate that personalization can also be evaluated through other objective performance outcomes, such as metabolic cost and interaction forces. The little differences observed in our study, therefore, emphasize the need to explore whether more permissive, trajectory‑free, or compliant controllers would allow personalized kinematic features to manifest more clearly in subjective experience and objective outcomes~\cite{shushtari_humanexoskeleton_2024}.

Future work should systematically investigate the aforementioned factors by testing alternative cuff designs and interface stiffness, comparing fully guided and assistive/impedance-based control modes, and implementing coupled prediction models that explicitly encode inter-joint dependencies. At the same time, expanding gait databases to include more diverse populations and walking speeds, exploring alternative prediction models will be essential. Ultimately, trajectory generation should not only aim for accurate prediction but also allow refinement (e.g., step length or width adjustments) based on individual user feedback. Engaging directly with the final users to gather and analyze their feedback on gait adjustments should be a central focus in future research.

\section{Conclusions}

In this within-subject user experience pilot study, we explored user perceptions of personalized versus standard and random gait patterns using an exoskeleton capable of multi-planar motion assistance, focusing on enjoyment, comfort, and naturalness. We developed and evaluated a comprehensive kinematic model for the exoskeleton control and regression-based predictive models generating personalized hip, knee, and pelvis trajectories from anthropometric, demographic, and walking speed data. Subjective evaluations revealed minimal differences between gait pattern conditions. We found, however, that participants generally rated the gait pattern experienced last as more comfortable and natural compared to the first one, regardless of the gait pattern enforced, suggesting adaptation to the experimental setup. These findings indicate that, in a small representation of unimpaired users, personalization of gait kinematics has a reduced short-term impact compared to adaptation to the robotic system and overall system ergonomics. 

\section*{Acknowledgments}
We would like to thank Prof. Robert Riener and his SMS lab at ETH Zurich, and Hocoma AG for their support and assistance during hardware and control development. We would also like to thank Prof. Herman van der Kooij for his support during the development of the predictive models. 
\bibliographystyle{IEEEtran}

\bibliography{biblio}

\end{document}

\title{
 Supplementary Materials \\ 
\begin{large} 
  Personalized Gait Patterns During Exoskeleton-Aided Training May Have Minimal Effect on User Experience. Insights from a Pilot Study
\end{large} }
\onecolumn

\maketitle
\vspace{-2cm}
\section{Kinematic Model}

% \item The angles \(\theta_1\) and \(\theta_2\) (with respect to the \(y_A\)-axis of frame \(\{A\}\)) can then be obtained by solving the following system:
% \begin{equation} \label{eq:17}
% \begin{matrix}
% \frac{l_c}{2} - l_1 \sin\theta_1 = -\frac{l_c}{2} + l_2 \sin\theta_2 \\
% p_{ext} + l_1 \cos\theta_1 = p_{int} + l_2 \cos\theta_2
% \end{matrix}
% \end{equation}
% where \(l_c\) is the distance between the two parallel motors' linear shafts, \(l_1\) and \(l_2\) are the link lengths of the parallel mechanism (as indicated in the paper in Fig.3).

% \item To determine the orientation of the thigh link, we consider the point \(M\), located in the middle of the thigh link's section and perpendicular to point \(B\). Its position with respect to the world frame \(\{O\}\) is:
% %by replacing it's location in Eq. (\ref{eq:21}), the translation of \(M\) respect to \(B\) is reached by Eq. (\ref{eq:25}). and the position of \(M\) in world frame is derived from Eq. (\ref{eq:26})
% \begin{equation} \label{eq:25}
% \mat{T}_M^B = \mat{R}_H^B
%  \begin{bmatrix}
% 0 \\ 
% l_n \\
% 0
% \end{bmatrix}
% \end{equation}
% \begin{equation} \label{eq:26}
% \mat{T}^O_M = \mat{R}_A^O  (\mat{T}^A_B+\mat{T}_M^B)
% \end{equation}

% where \(\mat{T}_B^O\) is obtained by replacing \(\mat{T}_H^A\) in Eq.20 (from the paper: $ \mat{T}^O_H = \mat{R}_A^O  \mat{T}^A_H $) with \(\mat{T}_B^A\) from $\mat{T}^A_B = \mat{T}_E^A + \mat{T}^E_B$, yielding to:
% \begin{equation} \label{eq:28}
% \mat{T}^O_B = \mat{R}_A^O  \mat{T}^A_B
% \end{equation}

% \end{itemize}
\subsection{Global description}
Fig.~\ref{Fig:2}.a represents the full kinematic model of the right leg.
\subfile{files/KineStructure}%

The \textbf{pelvis module} includes a pelvis plate, centered at point \(P\) in  Fig.~\ref{Fig:2}.a, which is attached to the user's pelvis and can undergo both translational and rotational motion with respect to frame \(\mathcal{F}\). This is achieved by a compliant spring-based connection between the fixed back of the pelvis module and the pelvis plate (see Wyss, 2019, for further details). The pelvis module functions as a series elastic actuator (SEA), with one actuated degree of freedom (P01-48x240 motor, NTI AG LinMot, Switzerland), enabling lateral movement relative to a fixed frame. To prevent hindering the natural lateral movement of the pelvis, the BWS is laterally actuated through a lead-screw mechanism parallel to the pelvis's actuated degree of freedom. Note that the other parts of the BWS remain unchanged with respect to their commercial version, i.e., allow for static and dynamic users' weight unloading. 

\textbf{Hip actuation} is achieved through a closed-chain mechanism driven by two linear actuators per leg (P01-48x240 motors, LinMot, Switzerland). The two prismatic actuators per leg are located on a plane that can rotate around the x-axis of the global frame \(\mathcal{O}\) (see Fig.~\ref{Fig:2}.a). 
This configuration enables control of both hip flexion/extension and ab-/adduction, while maintaining a rigid joint rotation. %
This is achieved through a parallel closed-chain mechanism connected to the outputs of two joints (denoted by $P_{int}$ and $P_{ext}$ in Fig.~\ref{Fig:2}.a) located at the distal end of the shafts of the hip prismatic actuators. On the other end, this closed-chain structure is attached to the thigh link of the orthosis via two perpendicular revolute joints ($E$ and $B$ in Fig.~\ref{Fig:2}.a). 

\begin{figure}[h!]
    \centering
    \includegraphics[width=0.5\columnwidth]{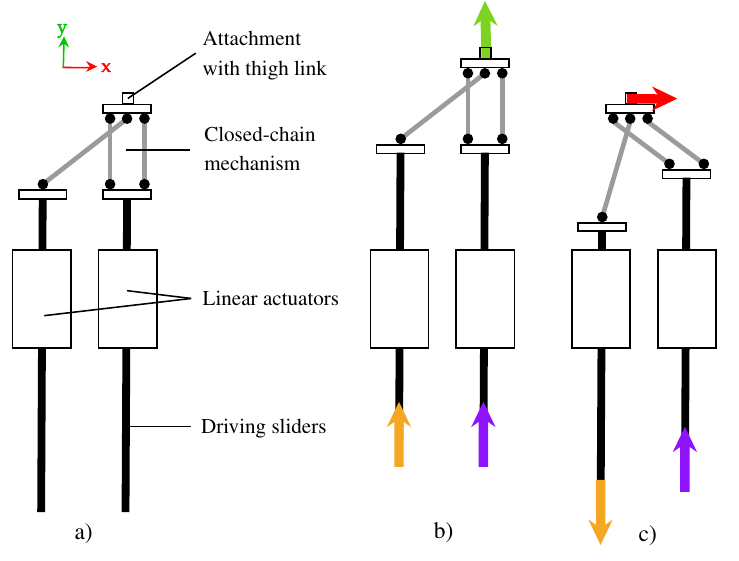}
    \caption{Top view of the right leg's closed-chain mechanism with the pelvis fixed in rotation, shown in different configurations: a) Neutral, b) Hip flexion/extension motion, and c) Hip ab-/adduction motion. Orange and purple arrows represent the movement of the linear actuators. Red and green arrows represent the resulting motion at the thigh.}
    \label{fig:hip-model}
    \vspace{-0.3cm}
\end{figure}

This mechanism converts the independent translational motions of the linear actuators into coordinated movements of the thigh link in the frontal and sagittal planes, as illustrated in Fig.~\ref{fig:hip-model}.

In particular, this configuration enforces hip flexion/extension movements when the two linear actuators move in the same direction (Fig.~\ref{fig:hip-model}.b), while ab-/adduction movements are produced by moving in opposite directions (Fig.~\ref{fig:hip-model}.c). By allowing the linear actuators to rotate about the x-axis of the passive joint $A$, the system avoids overconstraining the leg.

In Fig.~\ref{Fig:2}.a, the thigh rotational motion is represented by three rotational axes intersecting in $H$. Note that the hip internal/external rotation is mechanically constrained by the exoskeleton’s structure and, aside from small contributions from pelvis rotation, remains approximately fixed. 
%
%This mainly parallel actuation configuration offers several benefits over a serial approach like the FreeD, the most notable being a reduction in orthosis inertia, resulting in a more transparent system by design. % resulting  the robot a versatile system able to provide either stiff or compliant control.
%
For completeness, Fig.~\ref{Fig:2}.a also includes the \textbf{knee joint}, which is part of the original Lokomat\textsuperscript{\textregistered} device. The knee joint is located at the distal end of the thigh link, connected to the closed-chain structure via a revolute joint ($K$) and provided with a built-in force sensor. 
\subsection{Exoskeleton kinematic model}
\label{sec:kine}

We developed a closed-form analytical solution for the forward and inverse kinematics of the robot hip actuator module to be able to enforce personalized joint trajectories.
Since the mechanism does not incorporate any sensors to measure the rotations of the hip joint at $H$, we determined the direct and inverse mapping between the actuators' joint coordinates ($p_{int}$, $p_{ext}$, Fig.~\ref{Fig:2}.b) representing the shaft length of the two prismatic actuators of the hip module, and the space of hip joint coordinates ($\theta_{fl}$, $\theta_{ab}$, and $\theta_{ro}$, Fig.~\ref{Fig:2}.a), representing the hip flexion/extension, ab-/adduction, and internal/external rotation, respectively.

In the following paragraphs, we present the kinematic model and the process we followed for its definition, consisting of building the structure from both the perspective of the pelvis plate and the linear actuators to the spherical hip joint of the exoskeleton, and then closing the kinematic chain. 
Position and orientation of each joint will be expressed using homogeneous transformation matrices relative to the origin frame, \(\mathcal{O}\). In the following equations, \(\mathbf{T}_\mathcal{X}^\mathcal{Y}\) represents the translation vector from frame \(\mathcal{Y}\) to frame \(\mathcal{X}\), expressed in frame \(\mathcal{Y}\) coordinates, and \(\mathbf{R}_\mathcal{X}^\mathcal{Y}\) denotes the rotation matrix describing the orientation of frame \(\mathcal{X}\) relative to frame \(\mathcal{Y}\).

%\subsection{Kinematic loop-closure formulation}

The hip module of the exoskeleton forms a closed kinematic chain that can be expressed by closing two branches of the loop, either at the hip joint $H$ or at the upper hinge $E$ of the parallel mechanism (see Fig.~\ref{Fig:2}a). In the following, we derive the homogeneous positions of $H$ and $E$ from both closed-loop branches.

\subsubsection{Top branch (O $\rightarrow H \rightarrow M \rightarrow E$)}
The pelvis plate pose is defined by its measured rotations $\alpha,\beta,\gamma$, respect to fixed frame, around x,y and z axes respectively:
\begin{equation}
R^\mathcal{F}_\mathcal{P} = R_x(\alpha)R_y(\beta)R_z(\gamma), \qquad
T^\mathcal{O}_\mathcal{P} = T^\mathcal{O}_\mathcal{F} + T^\mathcal{F}_\mathcal{P} .
\end{equation}
The hip joint $H$, rigidly mounted on the pelvis plate, is located at a calibrated offset $T^\mathcal{P}_\mathcal{H}$ to be adaptable to accommodate different user sizes:
\begin{equation}
T^\mathcal{O}_\mathcal{H}\big|_{\text{top}} = T^\mathcal{O}_\mathcal{P} + R^\mathcal{F}_\mathcal{P} T^{P}_{H} .
\end{equation}
where the index \textit{top} indicates the propagation along the top branch.
The hip orientation in $O$ is given by
\begin{equation}
R^\mathcal{O}_\mathcal{H} = R^\mathcal{F}_\mathcal{P} R^\mathcal{P}_\mathcal{H}, \qquad
R^\mathcal{P}_\mathcal{H} = R_x(\theta_{fl})R_y(\theta_{ab})R_z(\theta_{ro}) .
\end{equation}
Propagating along the thigh yields
\begin{equation}
T^\mathcal{O}_\mathcal{E}\big|_{\text{top}} = T^\mathcal{O}_\mathcal{H}\big|_{\text{top}} + R^\mathcal{O}_\mathcal{H}
\begin{bmatrix}
0 \\ -l_n \\ -l_m
\end{bmatrix}.
\end{equation} 

\subsubsection{Bottom branch ($O \rightarrow E \rightarrow B \rightarrow H$)}
The actuator plane $A$ is rotated by $\theta_A$ about the $x$-axis:
\begin{equation}
R^\mathcal{O}_\mathcal{A} = R_x(\theta_A).
\end{equation}
The distal ends of the linear actuators are:
\begin{equation}
T^\mathcal{A}_{P_{\mathrm{ext}}} =
\begin{bmatrix}
\frac{l_c}{2} \\ p_{\mathrm{ext}} + d \\ 0
\end{bmatrix}, \quad
T^\mathcal{A}_{P_{\mathrm{int}}} =
\begin{bmatrix}
-\frac{l_c}{2} \\ p_{\mathrm{int}} + d \\ 0
\end{bmatrix}.
\end{equation}
With link orientations $R^\mathcal{A}_{P_{\mathrm{ext}}E} = R_z(\theta_1)$ and $R^\mathcal{A}_{P_{\mathrm{int}}E} = R_z(-\theta_2)$, the hinge $E$ satisfies:
\begin{equation}
T^\mathcal{A}_\mathcal{E} = T^\mathcal{A}_{P_{\mathrm{ext}}} + R^\mathcal{A}_{P_{\mathrm{ext}}E}
\begin{bmatrix}0\\l_1\\0\end{bmatrix}
= T^\mathcal{A}_{P_{\mathrm{int}}} + R^\mathcal{A}_{P_{\mathrm{int}}E}
\begin{bmatrix}0\\l_2\\0\end{bmatrix}.
\label{eq:TA_E}
\end{equation}

Multiplying Eq.~\ref{eq:TA_E} by \(\mathbf{R}_A^O\), the  position of the revolute joint \(E\) can be obtained:
\begin{equation}
T^\mathcal{O}_\mathcal{E}\big|_{\text{bot}} = R^\mathcal{O}_\mathcal{A} T^\mathcal{A}_\mathcal{E} =
\begin{bmatrix}
\frac{l_c}{2} - l_1 \sin \theta_1 \\
\cos \theta_A (p_\mathrm{ext} + d + l_1 \cos \theta_1) \\
\sin \theta_A (p_\mathrm{ext} + d + l_1 \cos \theta_1)
\end{bmatrix}.
\end{equation}
where the index \textit{bot} indicates the propagation through the bottom branch. Propagating further along the bottom branch:
\begin{align}
T^\mathcal{O}_\mathcal{B} &= T^\mathcal{O}_\mathcal{E}\big|_{\text{bot}} + R^\mathcal{O}_\mathcal{A} R_x(\theta_E)
\begin{bmatrix}0\\ l_n \\ 0\end{bmatrix}, \\
T^\mathcal{O}_\mathcal{H}\big|_{\text{bot}} &= T^\mathcal{O}_\mathcal{B} + R^\mathcal{O}_\mathcal{A} R_x(\theta_E) R_y(\theta_B)
\begin{bmatrix}0\\ 0\\ l_m\end{bmatrix}.
\end{align}

\subsubsection{Loop-closure conditions}
To solve for the unknown variables, the loop-closure equations are formulated by equating the two branches of the mechanism. 
The kinematic loop can be closed either at the hip joint ${H}$ or at the upper hinge ${E}$ of the parallel mechanism:
\begin{align}
&\text{Closure at $H$:} & T^\mathcal{O}_\mathcal{H}\big|_{\text{top}} &= T^\mathcal{O}_\mathcal{H}\big|_{\text{bot}} , \\
&\text{Closure at $E$:} & T^\mathcal{O}_\mathcal{E}\big|_{\text{top}} &= T^\mathcal{O}_\mathcal{E}\big|_{\text{bot}} .
\end{align}
Both systems of equations yield a valid solution to the closed-loop kinematics. 
The first formulation, closed at $H$, is particularly convenient for \textit{forward kinematics} (joint $\rightarrow$ actuator space), whereas the second, closed at $E$, is more convenient for \textit{inverse kinematics} (actuator $\rightarrow$ joint space).

The kinematic model of the mechanism was validated through an experiment performed by controlling the right leg of the exoskeleton in the execution of pre-recorded trajectories. Joint positions and angles obtained from the model were compared to those obtained from an OptiTrack\(\textsuperscript{\texttrademark}\) motion capture system. 

\subsection{Extra Details of Kinematic Model Calculations}
$\alpha$, $\beta$, and $\gamma$ denote the rotation angles of the pelvis frame $\mathcal{P}$ with respect to the fixed frame $\mathcal{F}$ about the $x$, $y$, and $z$ axes of the fixed frame, respectively, corresponding to roll, pitch, and yaw in an extrinsic X--Y--Z rotation sequence. The corresponding rotation matrices are:
\begin{equation}
R_x(\alpha) =
\begin{bmatrix}
1 & 0 & 0 \\
0 & \cos\alpha & -\sin\alpha \\
0 & \sin\alpha & \cos\alpha
\end{bmatrix},
\quad
R_y(\beta) =
\begin{bmatrix}
\cos\beta & 0 & \sin\beta \\
0 & 1 & 0 \\
-\sin\beta & 0 & \cos\beta
\end{bmatrix},
\quad
R_z(\gamma) =
\begin{bmatrix}
\cos\gamma & -\sin\gamma & 0 \\
\sin\gamma & \cos\gamma & 0 \\
0 & 0 & 1
\end{bmatrix}.
\end{equation}

The hip joint $H$ is modeled as a 3-DoF rotational joint. The joint coordinates
$\theta_{fl}$, $\theta_{ab}$, and $\theta_{ro}$ represent flexion/extension,
abduction/adduction, and internal/external rotation, respectively, defined as
extrinsic rotations about the $x$, $y$, and $z$ axes of the pelvis frame
$\mathcal{P}$.
The corresponding rotation matrices are
\begin{equation}
R_x(\theta_{fl}) =
\begin{bmatrix}
1 & 0 & 0 \\
0 & \cos\theta_{fl} & -\sin\theta_{fl} \\
0 & \sin\theta_{fl} & \cos\theta_{fl}
\end{bmatrix},
\quad
R_y(\theta_{ab}) =
\begin{bmatrix}
\cos\theta_{ab} & 0 & \sin\theta_{ab} \\
0 & 1 & 0 \\
-\sin\theta_{ab} & 0 & \cos\theta_{ab}
\end{bmatrix},
\quad
R_z(\theta_{ro}) =
\begin{bmatrix}
\cos\theta_{ro} & -\sin\theta_{ro} & 0 \\
\sin\theta_{ro} & \cos\theta_{ro} & 0 \\
0 & 0 & 1
\end{bmatrix}.
\end{equation}
% \begin{itemize}

\subsection{Validation of the kinematic model}
To validate the kinematic model of the mechanism, we performed an experiment with the right leg of the exoskeleton. After recording a healthy subject walking in the exoskeleton, while set in transparent mode, the obtained gait pattern was commanded to the exoskeleton (without, this time, any participant wearing it) in rigid position control mode.
The computed kinematics were compared against a ground-truth motion capture system.

During the execution of the pre-recorded trajectory, we recorded data from the device motor encoders and optical sensors, as well as from an OptiTrack\(^{TM}\) motion capture system.
The motion capture system consisted of four Prime\(^x\) 13W cameras, which measured the position and orientation of the key components: hip linear actuators, pelvis fixed frame, pelvis back plate, and thigh link.

\begin{figure}[H]
    \centering
    \includegraphics[width=0.3\columnwidth]{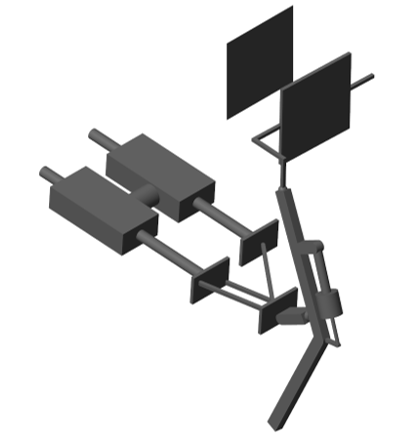}
    \caption{Screenshot of the mechanism Simscape model.}
    \label{fig:Simscape}
\end{figure}

To evaluate the \textbf{forward kinematics model}, we used the robot sensor data as inputs, including pelvis pose and prismatic joint positions recorded during the experiment. These inputs were fed into a Simulink/SimMechanics model of the system using MATLAB\textsuperscript{\textregistered} Simscape Multibody (R2021b, MathWorks, USA) to compute the corresponding hip joint angles, namely the flexion/extension, abduction/adduction, and internal/external rotation (see Fig.~\ref{fig:Simscape} for a screenshot). 

\begin{figure}[h!]
    \centering
  \includegraphics[width=0.4\columnwidth]{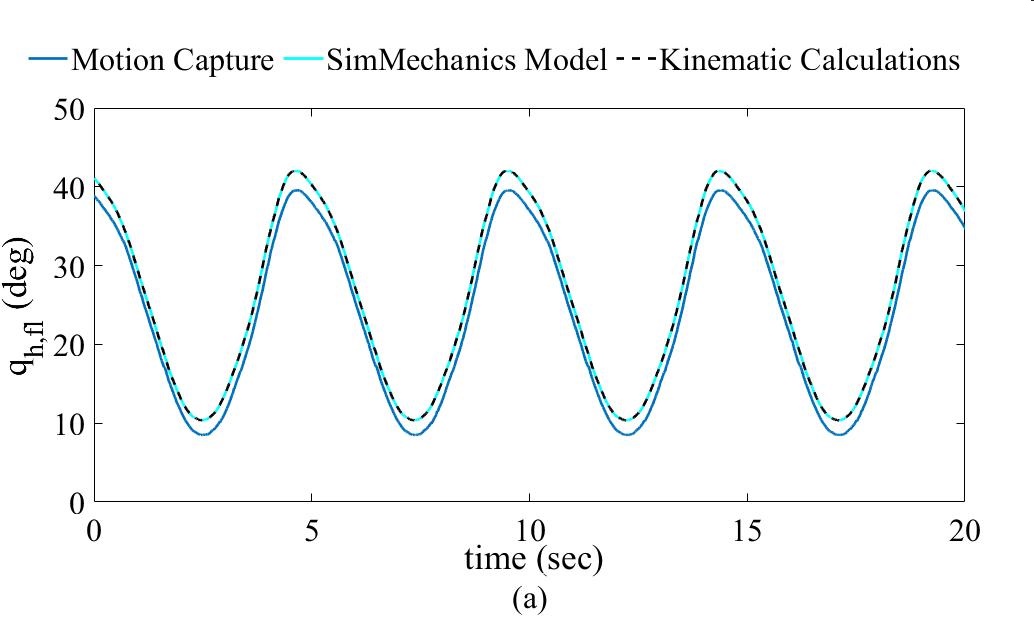}
\includegraphics[width=0.4\columnwidth]{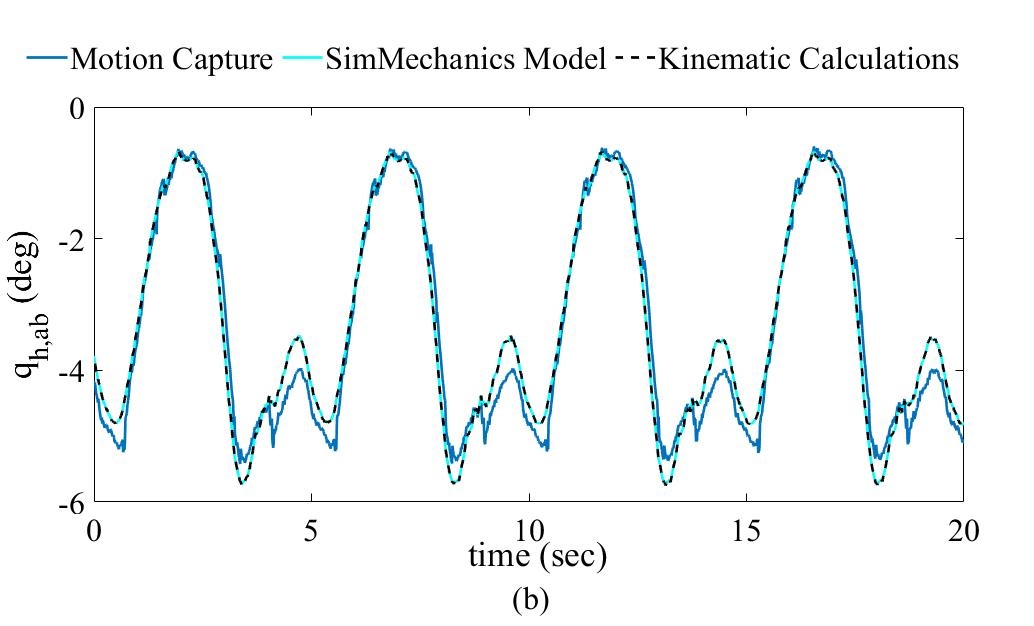}
    \includegraphics[width=0.4\columnwidth]{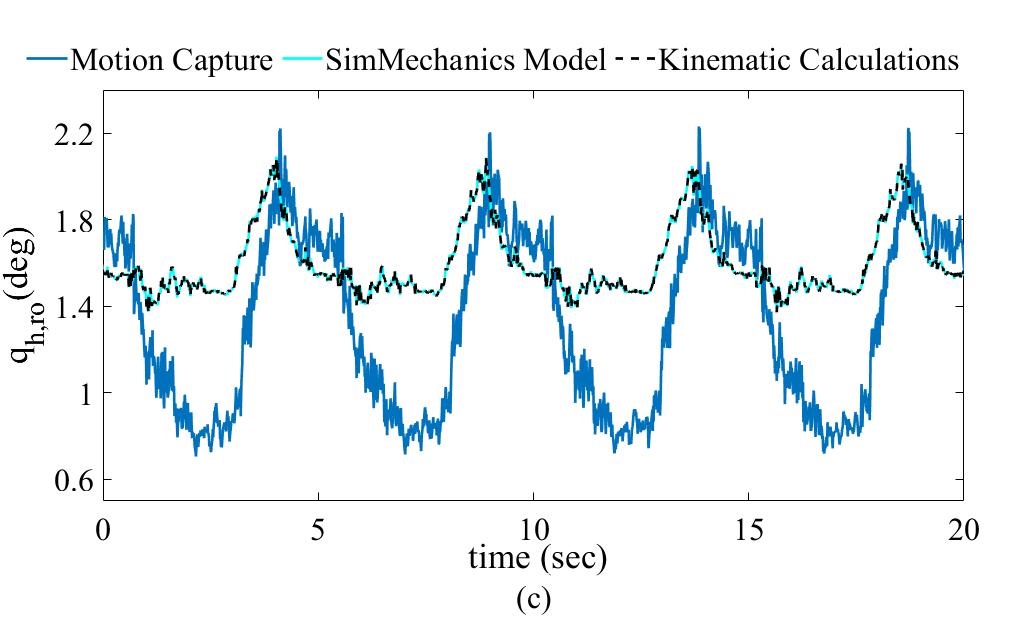}
\caption{Results from the forward kinematics validation. (a) Hip flexion/extension. (b) Hip abduction/adduction. (c) Hip rotation.}
\label{forward results}
\end{figure}

The simulated joint angles were then compared against those measured directly via the motion capture system to assess model accuracy (see Fig.\ref{forward results}).
Accuracy was assessed by computing maximum and root-mean-square errors (RMSE) between the recorded and simulation data.
The maximum error between the motion capture values and those calculated from the forward kinematics was \SI{2.97}{\degree} in hip flexion/extension,  
\SI{0.76}{\degree} in hip abduction/adduction and \SI{0.41}{\degree} in hip rotation. The RMSE was \SI{2.2}{\degree} in hip flexion/extension, \SI{0.32}{\degree} in hip abduction/adduction, and \SI{0.78}{\degree} in hip rotation.

\begin{figure}[h!]
    \centering
    \includegraphics[width=0.5\columnwidth]{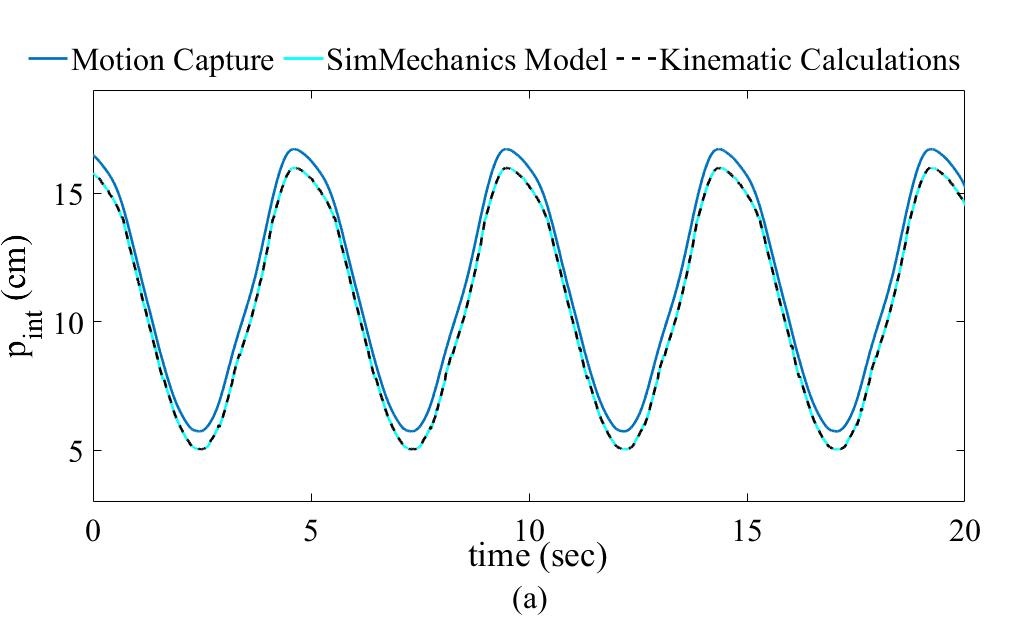}
    \includegraphics[width=0.5\columnwidth]{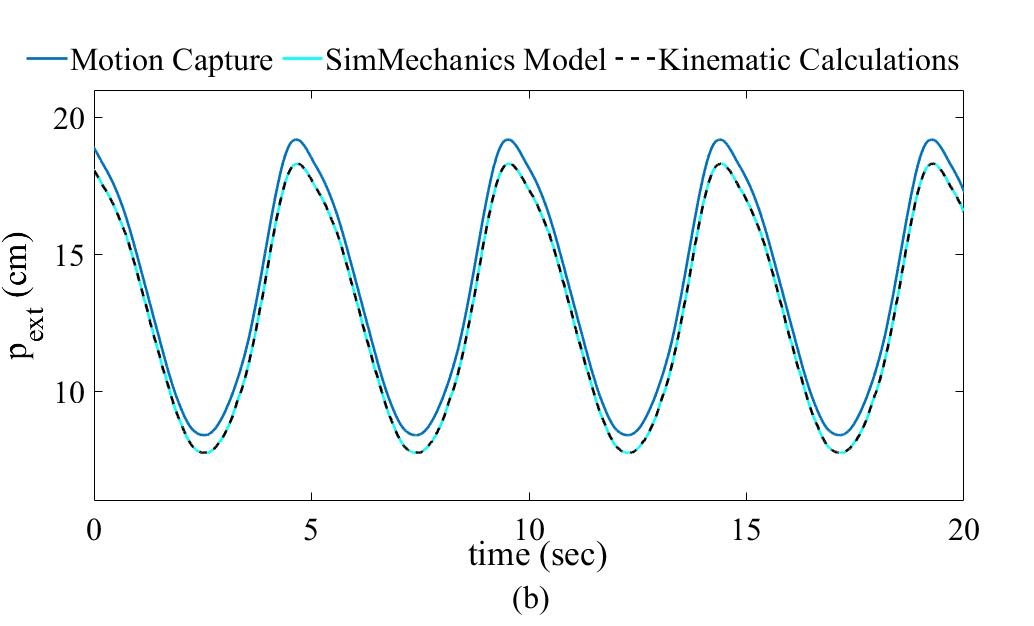}
    \caption{Results from the inverse kinematics. (a) Position of internal prismatic actuator. (b) Position of external prismatic actuator }
\label{inverse results}
\end{figure}

To validate the \textbf{inverse kinematics} mapping, we used hip joint angles measured via motion capture as inputs and computed the corresponding prismatic joint positions. These model outputs were then compared to the encoder readings of the linear actuators (see Fig.\ref{inverse results}). The maximum error for both internal and external actuators was below one centimeter, with root mean square errors (RMSE) of \SI{0.78}{\cm} and \SI{0.77}{\cm}, respectively.

%\stefano{what about the pelvis module? Was it kept fixed or free to move? especially in the inverse kinematics, we compute desired angles based on the assumption that the pelvis module is centered and fixed. If not, it is straightforward that we are going to have errors.}

\clearpage

\section{Regression Equations of Joint Trajectories}
\label{appendix-d}

The regression coefficients derived for the different joint trajectories and their corresponding key events are listed as follows: hip abduction/adduction in Table~\ref{tab:coefficients_hip_abduction}, hip flexion/extension in Table~\ref{tab:coefficients_hip_flexion}, knee flexion/extension in Table~\ref{tab:coefficients_knee_flexion}, and lateral pelvis movement in Table~\ref{tab:coefficients_lateral_pelvis}.

\begin{table*}[!ht]
\caption{Regression equations for the parameter values of the key events of hip abduction/adduction.}
\begin{tabularx}{\textwidth}{p{2.3cm}p{2.3cm}*{7}{X}}
\hline
\multicolumn{9}{l}{\textbf{Hip abduction/adduction}} \\
\hline
Key event & Parameter
& \begin{tabular}[c]{@{}l@{}}$\beta_c$ \\ (Intercept)\end{tabular}
& \begin{tabular}[c]{@{}l@{}}$\beta_v$ \\ (Speed)\end{tabular}
& \begin{tabular}[c]{@{}l@{}}$\beta_{v^2}$ \\ (Speed\textsuperscript{2})\end{tabular}
& \begin{tabular}[c]{@{}l@{}}$\beta_h$ \\ (Height)\end{tabular}
& \begin{tabular}[c]{@{}l@{}}$\beta_w$ \\ (Weight)\end{tabular}
& \begin{tabular}[c]{@{}l@{}}$\beta_a$ \\ (Age)\end{tabular}
& \begin{tabular}[c]{@{}l@{}}$\beta_s$ \\ (Gender)\end{tabular} \\
\hline
\multirow{4}{*}{Heel contact} & Timing ($t$) & 0 & - & - & - & - & - & - \\
             & Angle ($\theta$) & 0.0400 & - & - & - & - & - & - \\
             & Vel. ($\dot{\theta}$) & 0.0222 & - & 0.0075 & - & - & - & $-$0.0343 \\
             & Acc. ($\ddot{\theta}$) & 0.0390 & - & 0.0042 & - & - & - & $-$0.0182 \\
\hline
\multirow{4}{*}{Max. stance} & Timing ($t$) & 35.8245 & - & - & $-$10.0179 & - & - & 0.7653 \\
            & Angle ($\theta$) & 15.0676 & - & - & $-$8.2105 & 0.0568 & - & $-$1.3062 \\
            & Vel. ($\dot{\theta}$) & 0 & - & - & - & - & - & - \\
            & Acc. ($\ddot{\theta}$) & $-$0.0914 & - & - & - & - & 0.0006 & 0.0108 \\
\hline
\multirow{4}{*}{Max. $\dot{y}$ stance} & Timing ($t$) & 41.0000 & - & - & - & - & - & - \\
                    & Angle ($\theta$) & 17.5024 & - & - & $-$12.0096 & 0.0790 & - & $-$0.5726 \\
                    & Vel. ($\dot{\theta}$) & $-$0.0130 & - & - & - & - & - & - \\
                    & Acc. ($\ddot{\theta}$) & 0 & - & - & - & - & - & - \\
\hline
\multirow{4}{*}{Min. $\dot{y}$ swing} & Timing ($t$) & 58.3550 & - & - & - & - & - & - \\
                   & Angle ($\theta$) & $-$4.0730 & - & - & - & - & 0.0479 & - \\
                   & Vel. ($\dot{\theta}$) & $-$2.3595 & - & $-$0.0138 & 1.4393 & $-$0.0105 & 0.0034 & - \\
                   & Acc. ($\ddot{\theta}$) & 0 & - & - & - & - & - & - \\
\hline
\multirow{4}{*}{Min. swing} & Timing ($t$) & 74.1734 & $-$1.7955 & - & - & - & - & - \\
           & Angle ($\theta$) & $-$19.8236 & - & - & 8.9280 & $-$0.0598 & 0.0725 & - \\
           & Vel. ($\dot{\theta}$) & 0 & - & - & - & - & - & - \\
           & Acc. ($\ddot{\theta}$) & 0.2778 & - & - & $-$0.1289 & 0.0013 & $-$0.0008 & $-$0.0190 \\
\hline
\multirow{4}{*}{Max. $\dot{y}$ swing} & Timing ($t$) & 69.8878 & - & - & - & 0.0891 & 0.0331 & - \\
                   & Angle ($\theta$) & $-$5.2261 & - & - & - & - & 0.0444 & - \\
                   & Vel. ($\dot{\theta}$) & 0.4375 & $-$0.0366 & - & - & 0.0037 & $-$0.0030 & $-$0.0956 \\
                 & Acc. ($\ddot{\theta}$) & 0 & - & - & - & - & - & - \\
\hline
\end{tabularx}
\label{tab:coefficients_hip_abduction}
\vspace{0.0cm}\\
Note: ``-" indicates that a parameter has no significant impact on the regression model. The regression equations were generated using data from all 42 subjects. By default, heel strike timing is set to zero (\% of the gait cycle), as are the minimum and maximum values for joint angle and angular velocity.
\end{table*}

\begin{table*}[!ht]
\caption{Regression equations for the parameter values of the key events of hip flexion/extension.}
\begin{tabularx}{\textwidth}{p{2.3cm}p{2.3cm}*{7}{X}}
\hline
\multicolumn{9}{l}{\textbf{Hip flexion/extension}} \\
\hline
Key Event & Parameter 
& \begin{tabular}[c]{@{}l@{}}$\beta_c$ \\ (Intercept)\end{tabular} 
& \begin{tabular}[c]{@{}l@{}}$\beta_v$ \\ (Speed)\end{tabular} 
& \begin{tabular}[c]{@{}l@{}}$\beta_{v^2}$ \\ (Speed\textsuperscript{2})\end{tabular} 
& \begin{tabular}[c]{@{}l@{}}$\beta_h$ \\ (Height)\end{tabular} 
& \begin{tabular}[c]{@{}l@{}}$\beta_w$ \\ (Weight)\end{tabular} 
& \begin{tabular}[c]{@{}l@{}}$\beta_a$ \\ (Age)\end{tabular} 
& \begin{tabular}[c]{@{}l@{}}$\beta_s$ \\ (Gender)\end{tabular} \\
\hline
\multirow{4}{*}{Heel contact} & Timing ($t$) & 0 & - & - & - & - & - & - \\
             & Angle ($\theta$) & 24.4160 & - & - & - & - & 0.0830 & - \\
             & Vel. ($\dot{\theta}$) & $-$0.8169 & 0.1166 & - & 0.1915 & - & - & - \\
             & Acc. ($\ddot{\theta}$) & $-$0.3435 & 0.0775 & - & - & - & - & 0.0147 \\
\hline
\multirow{4}{*}{Max. stance\textsuperscript{a}} & Timing ($t$) & 15.6875 & - & - & $-$5.6864 & - & - & - \\
            & Angle ($\theta$) & 30.1610 & - & - & - & - & - & - \\
            & Vel. ($\dot{\theta}$) & 0 & - & - & - & - & - & - \\
            & Acc. ($\ddot{\theta}$) & $-$1.0905 & - & - & 0.7260 & $-$0.0043 & - & - \\
\hline
\multirow{4}{*}{$t$=50\% Stance} & Timing ($t$) & 31.4315 & $-$2.3754 & 0.3413 & - & - & - & - \\
              & Angle ($\theta$) & 5.4895 & - & - & - & - & 0.0868 & - \\
              & Vel. ($\dot{\theta}$) & $-$1.2312 & $-$0.1953 & - & 0.6511 & - & - & - \\
              & Acc. ($\ddot{\theta}$) & 0.0601 & - & $-$0.0017 & - & - & $-$0.0005 & - \\
\hline
\multirow{4}{*}{Min.} & Timing ($t$) & 63.4402 & $-$4.7863 & 0.6803 & - & - & - & - \\
     & Angle ($\theta$) & $-$4.5359 & - & - & - & - & - & 1.9210 \\
     & Vel. ($\dot{\theta}$) & 0 & - & - & - & - & - & - \\
     & Acc. ($\ddot{\theta}$) & 0.1171 & 0.0117 & - & - & - & - & - \\
\hline
\multirow{4}{*}{Max. $\dot{y}$ swing} & Timing ($t$) & 75.3750 & $-$9.3496 & 1.6596 & - & 0.0719 & - & $-$1.2761 \\
                 & Angle ($\theta$) & 8.4660 & - & - & - & - & - & - \\
                 & Vel. ($\dot{\theta}$) & 3.2209 & 0.0980 & - & $-$1.2215 & 0.0081 & $-$0.0033 & $-$0.0829 \\
                 & Acc. ($\ddot{\theta}$) & 0 & - & - & - & - & - & - \\
\hline
\multirow{4}{*}{Max. swing} & Timing ($t$) & 90.6226 & $-$1.0169 & - & - & - & 0.0474 & - \\
           & Angle ($\theta$) & 61.8406 & 2.0764 & - & $-$21.7837 & - & - & 1.4937 \\
           & Vel. ($\dot{\theta}$) & 0 & - & - & - & - & - & - \\
           & Acc. ($\ddot{\theta}$) & $-$0.1080 & - & - & - & - & - & - \\
\hline
\end{tabularx}
\label{tab:coefficients_hip_flexion}
\vspace{0.0cm} \\
\textsuperscript{a} This key event is excluded from the spline fitting procedure due to its relevance only for walking speeds exceeding 3.5 kph, whereas the developed regression model focuses on speeds up to 3.2 kph.
\end{table*}

\begin{table}[!ht]
\caption{Regression equations for the parameter values of the key events of knee flexion/extension.}
\begin{tabularx}{\textwidth}{p{2.3cm}p{2.3cm}*{7}{X}}
\hline
\multicolumn{9}{l}{\textbf{Knee flexion/extension}} \\
\hline
Key Event & Parameter 
& \begin{tabular}[c]{@{}l@{}}$\beta_c$ \\ (Intercept)\end{tabular} 
& \begin{tabular}[c]{@{}l@{}}$\beta_v$ \\ (Speed)\end{tabular} 
& \begin{tabular}[c]{@{}l@{}}$\beta_{v^2}$ \\ (Speed\textsuperscript{2})\end{tabular} 
& \begin{tabular}[c]{@{}l@{}}$\beta_h$ \\ (Height)\end{tabular} 
& \begin{tabular}[c]{@{}l@{}}$\beta_w$ \\ (Weight)\end{tabular} 
& \begin{tabular}[c]{@{}l@{}}$\beta_a$ \\ (Age)\end{tabular} 
& \begin{tabular}[c]{@{}l@{}}$\beta_s$ \\ (Gender)\end{tabular} \\
\hline
\multirow{4}{*}{Heel contact} & Timing ($t$) & 0 & - & - & - & - & - & - \\
             & Angle ($y$) & $-$18.4403 & - & $-$0.3887 & 10.1799 & - & 0.1908 & - \\
             & Vel. ($\dot{y}$) & $-$0.8710 & 0.7016 & $-$0.0875 & - & - & - & 0.0664 \\
             & Acc. ($\ddot{y}$) & $-$0.5245 & 0.2494 & - & - & - & - & 0.0490 \\
\hline
\multirow{4}{*}{Max. stance} & Timing ($t$) & 11.643 & - & - & - & - & - & - \\
            & Angle ($y$) & $-$2.2525 & - & 0.4534 & - & - & 0.2254 & 2.1273 \\
            & Vel. ($\dot{y}$) & 0 & - & - & - & - & - & - \\
            & Acc. ($\ddot{y}$) & $-$0.3389 & - & - & 0.3752 & $-$0.0074 & $-$0.0017 & 0.0743 \\
\hline
\multirow{4}{*}{Min. stance} & Timing ($t$) & 32.4633 & - & - & - & - & 0.1521 & 1.9152 \\
            & Angle ($y$) & 16.7077 & - & - & - & $-$0.2192 & - & 3.2653 \\
            & Vel. ($\dot{y}$) & 0 & - & - & - & - & - & - \\
            & Acc. ($\ddot{y}$) & 0.1067 & 0.0202 & - & $-$0.0980 & 0.0012 & - & - \\
\hline
\multirow{4}{*}{Max. $\dot{y}$ swing} & Timing ($t$) & 77.1875 & $-$7.9539 & 1.1427 & - & - & - & $-$0.4295 \\
                 & Angle ($y$) & 37.1826 & 1.4471 & - & - & $-$0.0980 & - & - \\
                 & Vel. ($\dot{y}$) & 3.2140 & - & - & - & 0.0061 & - & $-$0.1909 \\
                 & Acc. ($\ddot{y}$) & 0 & - & - & - & - & - & - \\
\hline
\multirow{4}{*}{Max. swing} & Timing ($t$) & 81.719 & $-$5.0453 & 0.8888 & - & - & - & - \\
           & Angle ($y$) & 52.1027 & 4.7823 & - & - & $-$0.1198 & - & - \\
           & Vel. ($\dot{y}$) & 0 & - & - & - & - & - & - \\
           & Acc. ($\ddot{y}$) & $-$0.9725 & 0.0768 & - & 0.1608 & - & 0.0014 & 0.0297 \\
\hline
\multirow{4}{*}{Min. $\dot{y}$ swing} & Timing ($t$) & 88.3116 & - & 0.1551 & - & - & - & - \\
                 & Angle ($y$) & 17.7490 & - & - & - & - & 0.1429 & - \\
                 & Vel. ($\dot{y}$) & $-$3.9013 & $-$0.4602 & - & - & 0.0143 & 0.0098 & - \\
                 & Acc. ($\ddot{y}$) & 0 & - & - & - & - & - & - \\
\hline
\end{tabularx}
\label{tab:coefficients_knee_flexion}
\end{table}

\begin{table*}[!ht]
\caption{Regression equations for the parameter values of the key events of lateral pelvis movement.}
\begin{tabularx}{\textwidth}{p{2.3cm}p{2.3cm}*{7}{X}}
\hline
\multicolumn{9}{l}{\textbf{Lateral pelvis movement}} \\
\hline
Key Event & Parameter 
& \begin{tabular}[c]{@{}l@{}}$\beta_c$ \\ (Intercept)\end{tabular} 
& \begin{tabular}[c]{@{}l@{}}$\beta_v$ \\ (Speed)\end{tabular} 
& \begin{tabular}[c]{@{}l@{}}$\beta_{v^2}$ \\ (Speed\textsuperscript{2})\end{tabular} 
& \begin{tabular}[c]{@{}l@{}}$\beta_h$ \\ (Height)\end{tabular} 
& \begin{tabular}[c]{@{}l@{}}$\beta_w$ \\ (Weight)\end{tabular} 
& \begin{tabular}[c]{@{}l@{}}$\beta_a$ \\ (Age)\end{tabular} 
& \begin{tabular}[c]{@{}l@{}}$\beta_s$ \\ (Gender)\end{tabular} \\
\hline
\multirow{4}{*}{Max. $\dot{y}$} & Timing ($t$) & 12.7882 & - & - & - & - & $-$0.0266 & - \\
           & Angle ($y$) & 6.9961 & - & - & - & - & $-$0.0735 & $-$1.7287 \\
           & Vel. ($\dot{y}$) & 7.1648 & $-$0.7182 & - & $-$2.0033 & 0.0210 & $-$0.0142 & - \\
           & Acc. ($\ddot{y}$) & 0 & - & - & - & - & - & - \\
\hline
\multirow{4}{*}{Max.} & Timing ($t$) & 26.1353 & - & - & 5.6096 & - & - & - \\
     & Angle ($y$) & 60.1065 & -9.2432 & - & - & - & - & - \\
     & Vel. ($\dot{y}$) & 0 & - & - & - & - & - & - \\
     & Acc. ($\ddot{y}$) & $-$0.1353 & - & 0.0031 & - & - & - & - \\
\hline
\multirow{4}{*}{Min. $\dot{y}$} & Timing ($t$) & 62.9191 & - & - & - & - & $-$0.0243 & - \\
           & Angle ($y$) & 4.2084 & - & - & - & $-$0.1631 & 0.0582 & 3.1179 \\
           & Vel. ($\dot{y}$) & $-$4.7921 & 0.7730 & - & - & - & - & - \\
           & Acc. ($\ddot{y}$) & 0 & - & - & - & - & - & - \\
\hline
\multirow{4}{*}{Min.} & Timing ($t$) & 75.8100 & - & - & 5.6217 & - & - & - \\
     & Angle ($y$) & $-$60.1309 & 9.2839 & - & - & - & - & - \\
     & Vel. ($\dot{y}$) & 0 & - & - & - & - & - & - \\
     & Acc. ($\ddot{y}$) & 0.0387 & - & $-$0.0018 & - & 0.0014 & - & -0.0212 \\
\hline
\end{tabularx}
\label{tab:coefficients_lateral_pelvis}
\end{table*}
\clearpage

\section{Quantitative data results}
The analysis of the exoskeleton's data recordings in terms of Mean Absolute Error (MAE) of actuator positions is reported in Table~\ref{tab:lmmResultsPositionError}.
\begin{table*}[h!]
\caption{Linear mixed model results for position error: \textit{Mean Absolute Error (MAE)} of the different actuators. BWS: Body Weight Support actuator.}
\label{tab:lmmResultsPositionError}
\begin{tabularx}{\textwidth}{X *{8}{l}}
% First question row
\toprule
\textbf{\multirow{2}{*}{Variable}} & \multicolumn{4}{l}{Pelvis Prismatic Actuator: MAE - Position (mm)} & \multicolumn{4}{l}{BWS Prismatic Actuator: MAE - Position (mm)} \\ 
\cmidrule(lr){2-5} \cmidrule(lr){6-9}
& \textbf{Estimate} ($\beta$) & \textbf{Std. Error} & \textbf{t value} & \textbf{p-value} & \textbf{Estimate} ($\beta$) & \textbf{Std. Error} & \textbf{t value} & \textbf{p-value} \\ 
\cmidrule(lr){1-9}
(Intercept) & 4.05 & 0.36 & 11.29 & \textbf{2.61E-11}*** & 2.54 & 0.21 & 11.96 & \textbf{1.48E-11}*** \\
Predicted pattern & 1.46 & 0.39 & 3.78 & \textbf{8.67E-04}** & 0.44 & 0.22 & 2.02 & 0.060 \\
Random pattern & 0.58 & 0.39 & 1.50 & 0.147 & 0.71 & 0.22 & 3.24 & \textbf{0.005}** \\
2nd Trial & 0.53 & 0.39 & 1.37 & 0.181 & 0.23 & 0.22 & 1.03 & 0.317 \\
3rd Trial & 0.14 & 0.39 & 0.37 & 0.713 & 0.05 & 0.22 & 0.22 & 0.827 \\
% Second question row
\midrule
\textbf{\multirow{2}{*}{Variable}} & \multicolumn{4}{l}{Right Outer Prismatic Actuator: MAE - Position (mm)} & \multicolumn{4}{l}{Right Inner Prismatic Actuator: MAE - Position (mm)} \\ 
\cmidrule(lr){2-5} \cmidrule(lr){6-9}
& \textbf{Estimate} ($\beta$) & \textbf{Std. Error} & \textbf{t value} & \textbf{p-value} & \textbf{Estimate} ($\beta$) & \textbf{Std. Error} & \textbf{t value} & \textbf{p-value} \\ 
\cmidrule(lr){1-9} 
(Intercept) & 8.35 & 0.15 & 56.41 & \textbf{8.77E-28}*** & 5.96 & 0.37 & 15.99 & \textbf{1.23E-14}*** \\
Predicted pattern & -0.53 & 0.14 & -3.85 & \textbf{0.001}** & -0.98 & 0.40 & -2.45 & \textbf{0.022}* \\
Random pattern & -1.13 & 0.14 & -8.17 & \textbf{4.20E-07}*** & -0.12 & 0.40 & -0.30 & 0.770 \\
2nd Trial & -0.20 & 0.14 & -1.48 & 0.157 & -0.92 & 0.40 & -2.30 & \textbf{0.030}* \\
3rd Trial & -0.18 & 0.14 & -1.32 & 0.205 & -0.32 & 0.40 & -0.81 & 0.425 \\
% Third question row
\midrule
\textbf{\multirow{2}{*}{Variable}} & \multicolumn{4}{l}{Left Outer Prismatic Actuator: MAE - Position (mm)} & \multicolumn{4}{l}{Left Inner Prismatic Actuator: MAE - Position (mm)} \\
\cmidrule(lr){2-5} \cmidrule(lr){6-9}
& \textbf{Estimate} ($\beta$) & \textbf{Std. Error} & \textbf{t value} & \textbf{p-value} & \textbf{Estimate} ($\beta$) & \textbf{Std. Error} & \textbf{t value} & \textbf{p-value} \\ 
\cmidrule(lr){1-9} 
(Intercept) & 8.48 & 0.15 & 58.43 & \textbf{3.26E-28}*** & 5.95 & 0.38 & 15.83 & \textbf{1.53E-14}*** \\
Predicted pattern & -0.52 & 0.14 & -3.84 & \textbf{0.001}** & -0.82 & 0.40 & -2.04 & 0.053 \\
Random pattern & -1.08 & 0.14 & -7.95 & \textbf{6.04E-07}*** & -0.13 & 0.40 & -0.32 & 0.753 \\
2nd Trial & -0.19 & 0.14 & -1.36 & 0.193 & -0.94 & 0.40 & -2.33 & \textbf{0.028}* \\
3rd Trial & -0.19 & 0.14 & -1.38 & 0.186 & -0.35 & 0.40 & -0.87 & 0.394 \\
% Fourth question row
\midrule
\textbf{\multirow{2}{*}{Variable}} & \multicolumn{4}{l}{Right Knee Revolute Actuator: MAE - Position (deg)} & \multicolumn{4}{l}{Left Knee Revolute Actuator: MAE - Position (deg)} \\
\cmidrule(lr){2-5} \cmidrule(lr){6-9}
& \textbf{Estimate} ($\beta$) & \textbf{Std. Error} & \textbf{t value} & \textbf{p-value} & \textbf{Estimate} ($\beta$) & \textbf{Std. Error} & \textbf{t value} & \textbf{p-value} \\ 
\cmidrule(lr){1-9} 
(Intercept) & 2.87 & 0.20 & 14.15 & \textbf{2.95E-12}*** & 3.06 & 0.27 & 11.46 & \textbf{4.91E-09}*** \\
Predicted pattern & -0.14 & 0.16 & -0.84 & 0.411 & -0.02 & 0.17 & -0.14 & 0.888 \\
Random pattern & 0.26 & 0.16 & 1.65 & 0.118 & 0.27 & 0.17 & 1.60 & 0.128 \\
2nd Trial & -0.47 & 0.16 & -2.96 & \textbf{0.009}** & -0.49 & 0.17 & -2.94 & \textbf{0.010}* \\
3rd Trial & -0.48 & 0.16 & -2.99 & \textbf{0.009}** & -0.25 & 0.17 & -1.48 & 0.157 \\
\bottomrule
\end{tabularx}

\vspace{0.15cm} 
\hspace{0.5cm}\raggedright *($p <0.05$), **($p <0.01$), ***($p <0.001$) 
\end{table*}

\newpage
\section{Questionnaires and Responses to Open-ended Questions}
\label{appendix:openEndedQuestions}

\subsection{User experience questionnaires}
\subsubsection{Questions}
Tab.\ref{tab:questionnaire} reports a summary of the questions asked per category assessing the user experience.
\begin{table}[h!]
\centering
\caption{Perception questionnaire items.}
\label{tab:questionnaire}
\begin{tabular}{|l|l|}
\hline
Category &
  Questions \\ \hline
\begin{tabular}[c]{@{}l@{}}Interest/ Enjoyment \\ (IMI)\end{tabular} &
  \begin{tabular}[c]{@{}l@{}}I enjoyed doing this activity very much\\ I thought this was a boring activity\\ I thought the task was very interesting\\ The activity was fun to do\end{tabular} \\ \hline
Passiveness &
  \begin{tabular}[c]{@{}l@{}}I remained passive during the activity\\ I resisted the movements of the exoskeleton\\ I allowed the exoskeleton to guide my lower limbs\\ I actively tried to control or influence the movements of the exoskeleton\end{tabular} \\ \hline
Comfort &
  \begin{tabular}[c]{@{}l@{}}How comfortable were the movements of the exoskeleton?\\ How would you rate the comfort at the cuffs during movement?\\ How comfortable were the movements in the hips?\\ How comfortable were the movements in the knees?\\ How comfortable were the movements in the ankles/feet?\\ I felt physical strain during the activity\\ I felt like I was secure in the device\end{tabular} \\ \hline
Naturalness &
  \begin{tabular}[c]{@{}l@{}}How natural did the movements of the exoskeleton feel?\\ How similar was the applied gait pattern compared to your own walking?\\ How smooth did the movements of the exoskeleton feel?\\ I felt the exoskeleton was pushing my limbs beyond my natural range.\end{tabular} \\\hline 
\end{tabular}
\end{table}

\subsubsection{Responses summary}
\begin{figure}[h!]
    \centering
    \includegraphics[width=0.75\linewidth]{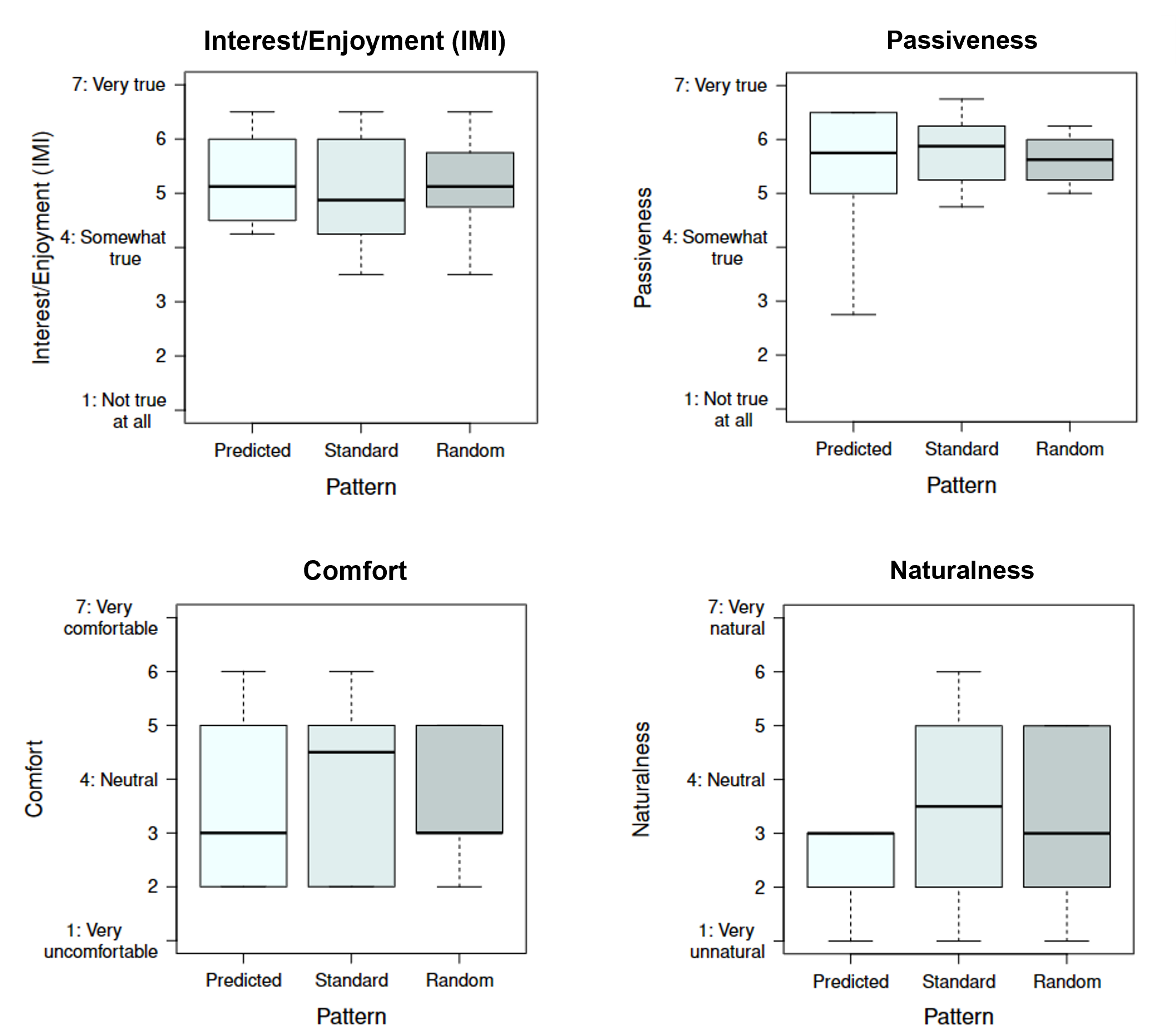}
    \caption{Scores attributed to the different categories of the questionnaire}
    \label{fig:scores}
\end{figure}

Figure \ref{fig:scores} presents the statistics of scores assigned to the various categories of the questionnaire.
\subsubsection{Linear Mixed Models}
Tab.\ref{tab:lmmResultsAll} reports the results of the Linear Mixed Models for the user-experience metrics.
\begin{table*}[h!]
\caption{Linear mixed model results for user experience metrics.}
\label{tab:lmmResultsAll}
\begin{tabularx}{\textwidth}{X *{8}{l}}
\toprule
\multicolumn{9}{l}{\textbf{Enjoyment and Passiveness}} \\
\midrule
 & \multicolumn{4}{l}{Enjoyment/Interest (IMI)\textsuperscript{2}} & \multicolumn{4}{l}{Passiveness\textsuperscript{2}} \\
\cmidrule(lr){2-5} \cmidrule(lr){6-9}
 & Estimate ($\beta$) & Std. Error & t value & p-value & Estimate ($\beta$) & Std. Error & t value & p-value \\
\cmidrule(lr){1-9}
(Intercept) & 5.01 & 0.30 & 16.58 & \textbf{5.28E-10}*** & 5.68 & 0.36 & 15.60 & \textbf{2.37E-14}*** \\
Personalized pattern & 0.20 & 0.15 & 1.34 & 0.199 & -0.30 & 0.36 & -0.84 & 0.413 \\
Random pattern & 0.13 & 0.15 & 0.85 & 0.409 & -0.29 & 0.36 & -0.81 & 0.429 \\
2nd Trial & 0.01 & 0.15 & 0.08 & 0.933 & 0.10 & 0.36 & 0.27 & 0.792 \\
3rd Trial & 0.02 & 0.15 & 0.12 & 0.907 & 0.10 & 0.36 & 0.28 & 0.781 \\

\midrule
\multicolumn{9}{l}{\textbf{Comfort}} \\
\midrule
& \multicolumn{4}{l}{Overall Comfort\textsuperscript{1}} & \multicolumn{4}{l}{Comfort at Cuffs\textsuperscript{1}} \\
\cmidrule(lr){2-5} \cmidrule(lr){6-9}
(Intercept) & 3.29 & 0.52 & 6.31 & \textbf{1.35E-06}*** & 3.81 & 0.49 & 7.82 & \textbf{2.88E-07}*** \\
Personalized pattern & -0.75 & 0.49 & -1.53 & 0.145 & 0.40 & 0.35 & 1.17 & 0.260 \\
Random pattern & -0.40 & 0.49 & -0.83 & 0.419 & 0.41 & 0.35 & 1.20 & 0.249 \\
2nd Trial & 0.96 & 0.49 & 1.97 & 0.067 & 0.14 & 0.35 & 0.41 & 0.688 \\
3rd Trial & 1.43 & 0.49 & 2.94 & \textbf{0.01}** & 0.10 & 0.35 & 0.29 & 0.774 \\

\midrule
 & \multicolumn{4}{l}{Comfort at Hips\textsuperscript{1}} & \multicolumn{4}{l}{Comfort at Knees\textsuperscript{1}} \\
\cmidrule(lr){2-5} \cmidrule(lr){6-9}
(Intercept) & 4.39 & 0.63 & 7.02 & \textbf{2.42E-07}*** & 5.16 & 0.57 & 9.00 & \textbf{2.72E-09}*** \\
Personalized pattern & -0.26 & 0.58 & -0.45 & 0.655 & -0.15 & 0.53 & -0.29 & 0.779 \\
Random pattern & -0.04 & 0.58 & -0.07 & 0.945 & -0.55 & 0.53 & -1.03 & 0.320 \\
2nd Trial & 0.60 & 0.58 & 1.03 & 0.317 & -0.45 & 0.53 & -0.86 & 0.405 \\
3rd Trial & 0.22 & 0.58 & 0.38 & 0.705 & 0.06 & 0.53 & 0.11 & 0.911 \\

\midrule
 & \multicolumn{4}{l}{Comfort at Ankles/Feet\textsuperscript{1}} & \multicolumn{4}{l}{Physical Strain\textsuperscript{2}} \\
\cmidrule(lr){2-5} \cmidrule(lr){6-9}
(Intercept) & 4.05 & 0.53 & 7.64 & \textbf{6.37E-08}*** & 2.10 & 0.43 & 4.86 & \textbf{6.14E-05}*** \\
Personalized pattern & -0.44 & 0.47 & -0.94 & 0.362 & 1.07 & 0.37 & 2.86 & \textbf{0.011}* \\
Random pattern & -0.28 & 0.47 & -0.60 & 0.559 & 0.44 & 0.37 & 1.19 & 0.252 \\
2nd Trial & 0.17 & 0.47 & 0.36 & 0.722 & 0.44 & 0.37 & 1.19 & 0.252 \\
3rd Trial & 0.62 & 0.47 & 1.30 & 0.212 & -0.26 & 0.37 & -0.70 & 0.492 \\

\midrule
 & \multicolumn{4}{l}{Sense of Security in Device\textsuperscript{2}} \\
\cmidrule(lr){2-5}
(Intercept) & 5.15 & 0.45 & 11.44 & \textbf{8.22E-10}*** \\
Personalized pattern & -0.13 & 0.32 & -0.41 & 0.687 \\
Random pattern & 0.22 & 0.32 & 0.69 & 0.497 \\
2nd Trial & 0.22 & 0.32 & 0.69 & 0.497 \\
3rd Trial & 0.54 & 0.32 & 1.67 & 0.114 \\

\midrule
\multicolumn{9}{l}{\textbf{Naturalness}} \\
\midrule
 & \multicolumn{4}{l}{Naturalness of Movements\textsuperscript{1}} & \multicolumn{4}{l}{Similarity to Own Way of Walking\textsuperscript{1}} \\
\cmidrule(lr){2-5} \cmidrule(lr){6-9}
(Intercept) & 2.68 & 0.52 & 5.20 & \textbf{2.25E-05}*** & 2.56 & 0.51 & 4.98 & \textbf{4.01E-05}*** \\
Personalized pattern & -0.71 & 0.49 & -1.45 & 0.165 & -0.76 & 0.51 & -1.50 & 0.154 \\
Random pattern & -0.17 & 0.49 & -0.35 & 0.728 & -0.58 & 0.51 & -1.14 & 0.272 \\
2nd Trial & 1.28 & 0.49 & 2.64 & 0.018 & 1.24 & 0.51 & 2.45 & 0.026 \\
3rd Trial & 1.35 & 0.49 & 2.79 & \textbf{0.013}** & 1.82 & 0.51 & 3.59 & \textbf{0.002}** \\

\midrule
 & \multicolumn{4}{l}{Smoothness of Movements\textsuperscript{1}} & \multicolumn{4}{l}{Limbs Pushed Beyond Natural Range\textsuperscript{2}} \\
\cmidrule(lr){2-5} \cmidrule(lr){6-9}
(Intercept) & 4.61 & 0.40 & 11.42 & \textbf{2.09E-10}*** & 2.00 & 0.40 & 5.04 & \textbf{5.21E-05}*** \\
Personalized pattern & 0.16 & 0.31 & 0.52 & 0.614 & -0.09 & 0.32 & -0.29 & 0.777 \\
Random pattern & 0.11 & 0.31 & 0.35 & 0.728 & -0.42 & 0.32 & -1.35 & 0.197 \\
2nd Trial & 0.11 & 0.31 & 0.35 & 0.728 & -0.24 & 0.32 & -0.77 & 0.453 \\
3rd Trial & 0.49 & 0.31 & 1.58 & 0.134 & -0.33 & 0.32 & -1.06 & 0.306 \\
\bottomrule
\end{tabularx}

\vspace{0.15cm} 
\hspace{0.15cm}
  \raggedright  *($p <0.05$), **($p <0.01$), ***($p <0.001$)\\
 \textsuperscript{1} 7-point Likert Scale: (1) Very unnatural/uncomfortable/dissimilar - (4) Neutral - (7) Very natural/comfortable/similar \\
 \textsuperscript{2} 7-point Likert Scale: (1) Not true at all - (4) Somewhat true - (7) Very true 
\end{table*}

\subsection{Open ended questions}
The answers of the open-ended questions are organized into the tables below: Table~\ref{tab:openEndedResultsQ1} contains responses related to discomfort and unusual sensations; Table~\ref{tab:openEndedResultsQ2} contains responses related to aspects influencing comfort or discomfort; and Table~\ref{tab:openEndedResultsQ3} includes additional thoughts or comments provided by participants.

\begin{table*}[h!]
\caption{Participant responses regarding the open-ended question about discomfort and unusual sensations, sorted by gait patterns (standard, personalized, random).}
\label{tab:openEndedResultsQ1}
\begin{tabularx}{\textwidth}{|c|X|X|X|}
\hline
\textbf{\multirow{2}{*}{Participant}} & 
\multicolumn{3}{c|}{\begin{tabular}{c} \textbf{Open-ended question:}\\ \multicolumn{1}{p{15cm}}{\centering Did you experience any discomfort in terms of physical strain or unusual sensations during the walking session with the robot? \\ If so, please specify the level of discomfort as well as the areas of discomfort, e.g., on which cuffs, joints, or other body parts.}\end{tabular}} \\ \cline{2-4}
& \textbf{Standard} & \textbf{Personalized} & \textbf{Random} \\ \hline
1 & ``I felt like there was some weight on my feet which made it difficult to remain passive. Discomfort level: 6/10." & ``I felt discomfort with how much weight was applied on my feet. Because of this it was hard to remain passive. Discomfort level: 7/10." & ``I felt some discomfort at the hips (6/10)." \\ \hline
2 & ``Cuffs at the ankles still uncomfortable as [during the random pattern]. Much less strain to keep foot position" & ``Very similar to [standard pattern]." & ``Yes, the machine was too low while walking and my feet did not have room to go back. At the same time if the machine was higher the ankle cuffs pressed too much against my calf. Also having to keep my feet as horizontal as possible was tiring." \\ \hline
3 & ``I still felt strain around the ankle cuffs seemingly pushing my calf muscles up, although it was less compared to the last trial. I also felt some chafing around the hip bone this time; minor but there" & ``Ankle cuffs keep pushing the muscles of my calves up (on and off). This creates a kind of cramped feeling sometimes. Also by nature of the device it feels a little as if the space between my thighs is larger than it normally would be (but this may be due to the fittings there)." & ``This pattern resulted in the least chafing for me, if none, but the experiment progressed, the excessive hip swaying began to hurt at the hip bone. Also the ankles started to chafe again as everything progressed." \\ \hline
4 & - & - & ``Yes, a slight discomfort in the knees." \\ \hline
5 & ``Yes, particularly uncomfortable on the hips." & ``Slight discomfort at ankles (especially left)." & ``Light discomfort on calf cuff. Medium discomfort at ankle and knee." \\ \hline
6 & - & ``My muscles got a bit sore since you relax your muscles except for the calves because of the ankle movements." & ``No, this was much better compared to the first two gait patterns. However, this could also be because I got used to it." \\ \hline
7 & - & - & - \\ \hline
8 & ``Ankle cuffs a bit uncomfortable." & ``Ankle cuffs uncomfortable." & ``Right ankle cuff was sometimes a bit uncomfortable." \\ \hline
9 & ``I felt slight discomfort in my ankles when walking with the exoskeleton. The joints were moved in an unnatural position." & ``I experienced some discomfort in my legs around the cuffs, because my legs were pushed beyond the natural range of motion." & ``No physical strain." \\ \hline
10 & - & ``Mild discomfort on the outer part of my feet when landing after every step." & - \\ \hline
\end{tabularx}
\end{table*}

\begin{table*}[h!]
\caption{Participant responses regarding the open-ended question about aspects promoting comfort or discomfort, sorted by gait patterns (standard, personalized, random).}
\label{tab:openEndedResultsQ2}
\begin{tabularx}{\textwidth}{|c|X|X|X|}
\hline
\textbf{\multirow{2}{*}{Participant}} & 
\multicolumn{3}{c|}{\begin{tabular}{c} \textbf{Open-ended question:}\\ \multicolumn{1}{p{15cm}}{\centering Were there any specific aspects of the gait pattern that contributed to your comfort or discomfort? \\ If so, please specify. For example, consider aspects such as an unusual movement of a particular joint, a small/ large range of motion, jerking of the trajectory, or unsymmetrical gait pattern between left and right leg.}\end{tabular}} \\ \cline{2-4}
& \textbf{Standard} & \textbf{Personalized} & \textbf{Random} \\ \hline
1 & ``My feet were moving too close to each other which was unnatural for me." & ``Again I felt like my feet were too close to each other during the trial." & ``I felt like my feet were a bit too close to each other when walking, which was unnatural for me." \\ \hline
2 & ``Acceleration still a bit abrupt." (referred to random pattern) & ``I felt the accelerations were less abrupt compared to the other two conditions." & ``I felt the forward acceleration phase was unnaturally fast." \\ \hline
3 & ``My hips were still swaying quite far sideways. This time it was less problematic because the movement was less so and my body partially countered the weight distributions during. This made it feel less like my body was being thrown left to right by the hips." & ``Too felt my hips were swaying outside the range my legs (or really where my feet are positioned on the ground), creating a sensation as if I would fall [if] I [have] to walk like this outside of the device." & ``There was once again excessive hip swaying, the worst of the three conditions." \\ \hline
4 & - & - & ``Not sure about the source but my knees felt rather stiff." \\ \hline
5 & ``The trajectory was moving my waist left and right, making the hip movement less pleasant and natural." & ``My left leg was pushed a bit towards the right, making the left ankle acting a bit shaky at the end of the motion." & "Yes, my legs were positioned more forward than their natural position, this implied better movement for the hip but worse for the knee. It was difficult to raise the tip of the foot enough." \\ \hline
6 & ``The gait pattern was somewhat counterintuitive since it felt like I wasn't walking in a straight line." & ``The lower limb movement was ok, but my feet were towards the centre. Also, hips are moved too much towards the outside $\rightarrow$ uncomfortable!" & ``Now, I know why it was uncomfortable; feet are too distant from each other, so you have to compensate with your hips $\rightarrow$ uncomfortable." \\ \hline
7 & ``This time my legs seemed to move from the outside to the inside when extending the leg to the front. This felt more similar to my usual gait." & ``I felt like the legs were sometimes moving from the inside to the outside during leg extension to the front. Depending on if I was leaning more to the right or left it was unsymmetrical." & ``The hips were moving less from side to side in my opinion this made it more comfortable. But the legs were moving more to the outside when extending the leg. This was a bit unnatural." \\ \hline
8 & ``Also a lot of lateral hip movement." & ``Also lateral hip movement [uncomfortable]." & ``Too much lateral movement of my hips felt funny." \\ \hline
9 & ``The gait pattern moved my feet in an unnatural movement at the heel strike." & ``My discomfort was high because there was an unnatural motion for my legs to walk with." & ``Comfort in my ankles was higher due to smaller movement/ slower jerking motions compared to 1st trial [standard pattern]." \\ \hline
10 & ``I felt this time the steps were wider and the trajectory the foot followed from start to end of the step was straighter, which made it more similar to my gait." & ``The trajectory of my knees/ankles felt unnatural because it went in and out before landing the step." & ``Ankles were pushed inwards, which made it feel unnatural. Also, knee movement had some overshoot and then retrieved a bit before landing the feet." \\ \hline
\end{tabularx}
\end{table*}

\begin{table*}[h!]
\caption{Participant responses regarding the open-ended question on additional thoughts or comments, sorted by gait patterns (standard, personalized, random).}
\label{tab:openEndedResultsQ3}
\begin{tabularx}{\textwidth}{|c|X|X|X|}
\hline
\textbf{\multirow{2}{*}{Participant}} & 
\multicolumn{3}{c|}{\begin{tabular}{c} \textbf{Open-ended question:}\\ \multicolumn{1}{p{15cm}}{\centering Would you like to share any other thoughts or comments?}\end{tabular}} \\ \cline{2-4}
& \textbf{Standard} & \textbf{Personalized} & \textbf{Random} \\ \hline
1 & - & - & - \\ \hline
2 & - & - & - \\ \hline
3 & - & - & ``1) Despite the points above, the legs movement \& time felt the most natural for this [trial]. The legs felt too spread \& the hip swaying felt like I would fall. The legs movements overall felt quite natural somehow.\newline 2) Sometimes the machine movements felt a little odd during this trial, as if it would speed up very briefly, it did this maybe twice so it might not be anything (just an observation)." \\ \hline
4 & ``The hip movement felt a bit unusual, perhaps slightly exaggerated." & ``It wasn't discomfort per se but the hip movement felt over-exaggerated." & - \\ \hline
5 & - & ``I felt that my weight was placed more on the right leg, like a 60-40 subdivision of the weight." & - \\ \hline
6 & - & ``Because of the unnatural movements, I really had to focus on what I was doing instead of letting go and naturally walk." & - \\ \hline
7 & - & - & - \\ \hline
8 & - & - & - \\ \hline
9 & ``Rest of gait pattern as moved natural." & ``I think the unnatural movements were contributing to the strain/discomfort around the cuffs, but I am not sure." & - \\ \hline
10 & - & - & - \\ \hline
\end{tabularx}
\end{table*}